\PassOptionsToPackage{table}{xcolor}
\documentclass{article}
\usepackage{iclr2027_conference,times}

\usepackage{amsmath,amsfonts,amssymb,bm}

\def\eqref#1{equation~\ref{#1}}
\def\1{\bm{1}}

\DeclareMathAlphabet{\mathsfit}{\encodingdefault}{\sfdefault}{m}{sl}
\SetMathAlphabet{\mathsfit}{bold}{\encodingdefault}{\sfdefault}{bx}{n}

\usepackage{booktabs}
\usepackage{graphicx}
\usepackage{algorithm}
\usepackage{algorithmic}
\usepackage{microtype}
\usepackage{multirow}
\usepackage{placeins}
\usepackage{caption}
\usepackage{xcolor}
\usepackage{tcolorbox}
\newtcolorbox{trajsys}{colback=gray!6,colframe=gray!55,boxrule=0.4pt,arc=1pt,
  left=4pt,right=4pt,top=2pt,bottom=2pt,before skip=4pt,after skip=2pt,
  title={\scriptsize\bfseries SYSTEM PROMPT},fonttitle=\scriptsize,
  coltitle=black,colbacktitle=gray!18,boxrule=0.4pt}
\newtcolorbox{trajusr}{colback=blue!5,colframe=blue!40,boxrule=0.4pt,arc=1pt,
  left=4pt,right=4pt,top=2pt,bottom=2pt,before skip=4pt,after skip=2pt,
  title={\scriptsize\bfseries USER PROMPT},fonttitle=\scriptsize,
  coltitle=black,colbacktitle=blue!14}
\newtcolorbox{trajasst}{colback=green!5,colframe=green!45!black,boxrule=0.4pt,
  arc=1pt,left=4pt,right=4pt,top=2pt,bottom=2pt,before skip=4pt,after skip=2pt,
  title={\scriptsize\bfseries MODEL RESPONSE},fonttitle=\scriptsize,
  coltitle=black,colbacktitle=green!16}
\newtcolorbox{trajcall}{colback=orange!7,colframe=orange!65!black,boxrule=0.4pt,
  arc=1pt,left=4pt,right=4pt,top=2pt,bottom=2pt,before skip=4pt,after skip=2pt,
  title={\scriptsize\bfseries TOOL CALL},fonttitle=\scriptsize,
  coltitle=black,colbacktitle=orange!22}
\newtcolorbox{trajresp}{colback=violet!5,colframe=violet!50,boxrule=0.4pt,
  arc=1pt,left=4pt,right=4pt,top=2pt,bottom=2pt,before skip=4pt,after skip=2pt,
  title={\scriptsize\bfseries TOOL RESPONSE},fonttitle=\scriptsize,
  coltitle=black,colbacktitle=violet!16}

\newtcolorbox{trajenv}{colback=violet!5,colframe=violet!50,boxrule=0.4pt,
  arc=1pt,left=4pt,right=4pt,top=2pt,bottom=2pt,before skip=4pt,after skip=2pt,
  title={\scriptsize\bfseries ENVIRONMENT RESPONSE},fonttitle=\scriptsize,
  coltitle=black,colbacktitle=violet!16}

\newtcolorbox{findingbox}{colback=blue!4,colframe=blue!30!gray,boxrule=0.5pt,arc=2pt,left=6pt,right=6pt,top=4pt,bottom=4pt,before skip=9pt,after skip=0pt}
\usepackage{fvextra}
\usepackage{hyperref}
\usepackage{url}

\title{
  WHALE: A Simple Recipe for Joint Harness--Weight Optimization
}

\author{Haechan Kim$^{1,2}$, Yoonho Lee$^{3}$, Gisang Lee$^{1}$, Chelsea Finn$^{3}$, Kangwook Lee$^{1}$ \\
$^{1}$KRAFTON, $^{2}$KAIST, $^{3}$Stanford University \\
\texttt{kim.haechan@kaist.ac.kr}}

\newcommand{\method}{WHALE}
\newcommand{\meanatk}{\ensuremath{\operatorname{mean@8}}}

\newif\ifarxiv
\arxivtrue
\ifarxiv
  \iclrfinalcopy
  \newcommand{\atapp}[1]{}
  \newcommand{\herowidth}{0.95\textwidth}
\else
  \newcommand{\atapp}[1]{ in Appendix~\ref{#1}}
  \newcommand{\herowidth}{0.84\textwidth}
\fi

\begin{document}

% one paragraph gap everywhere: run-in headings use \parskip like ordinary paragraphs
\makeatletter
\def\paragraph{\@startsection{paragraph}{4}{\z@}{0pt}{-1em}{\normalsize\bf}}
\makeatother

% tighter float and caption spacing
\setlength{\textfloatsep}{6pt plus 2pt minus 2pt}
\setlength{\floatsep}{8pt plus 2pt minus 2pt}
\setlength{\intextsep}{8pt plus 2pt minus 2pt}
\setlength{\abovecaptionskip}{6pt}
\setlength{\belowcaptionskip}{0pt}

\maketitle
% the arXiv preprint carries no venue banner in the running head
\ifarxiv\lhead{}\fi

\begin{abstract}
Agent performance depends jointly on the model parameters and the executable harness code that manages context and control flow.
Optimizing either component in isolation can leave the system bottlenecked by its frozen counterpart: weight updates can change which harness is effective, while harness updates can change which model capabilities are exposed.
Existing joint-adaptation methods optimize weights and textual prompts but leave the broader harness fixed.
We propose \textbf{W}eight-\textbf{H}arness \textbf{A}lternating \textbf{LE}arning (\textbf{\method{}}), a simple recipe that alternates two phases: updating the model under the current harness, then searching for a better harness under the updated model.
We instantiate these two phases with online rejection-sampling fine-tuning and Meta-Harness~\citep{lee2026metaharness}, respectively.
When to switch is a key design choice: to separate real improvements from noise without over-optimizing against a changing counterpart, \method{} uses either fixed phase durations or an adaptive patience rule over training signals.
Using Qwen3.5-2B/4B agents across three domains (search question answering, mathematical reasoning, and chess puzzles), \method{} outperforms weight-only, harness-only, and Fast--Slow Training~\citep{tiwari2026fst} by 4.15--24.38 percentage points in best \meanatk{} accuracy.
Either component can be the bottleneck: harness search matches peak weight-only accuracy with far fewer rollouts in SearchQA, but improves math accuracy only after a weight update.
Small interleaved updates also outperform stagewise weight-then-harness optimization in accuracy and rollout cost.
\ifarxiv\ The code is available at \url{https://github.com/krafton-ai/WHALE}.\fi
\end{abstract}

% A controlled schedule study shows that small alternating steps reach higher accuracy with fewer rollouts than stagewise optimization. We find that phases that are too short can select noisy updates, whereas longer phases lose accuracy, consistent with over-optimization against a frozen counterpart.
% Adaptive \method{} stops each phase using training-signal patience, exceeding the main fixed schedule in both studied domains without selecting fixed per-cycle budgets.
% We further run a controlled, budget-matched study of the alternation schedule, spanning a stagewise schedule, five fixed alternating schedules, and an adaptive schedule. Together the experiments yield six findings: (1) \method{} consistently outperforms single-component updates across domains; (2) domains divide into harness-dominant and model-dominant regimes, with harness search the more rollout-efficient lever for capabilities that both operators can improve; (3) a small weight update can unlock an otherwise ineffective harness search; (4) one large pass per component is dominated by small alternating steps in both accuracy and rollout cost; (5) the same small-step schedule is strongest in both regimes, since per-cycle budgets must lie between noisy and over-optimized choices; and (6) a per-phase patience rule on training signals finds this interval automatically, and the resulting adaptive \method{} matches or exceeds the fixed schedules.
%%YL.8.26: 6 points was a lot for an abstract

\begin{figure*}[!ht]
\centering
\vspace{-1mm}
\includegraphics[width=\herowidth]{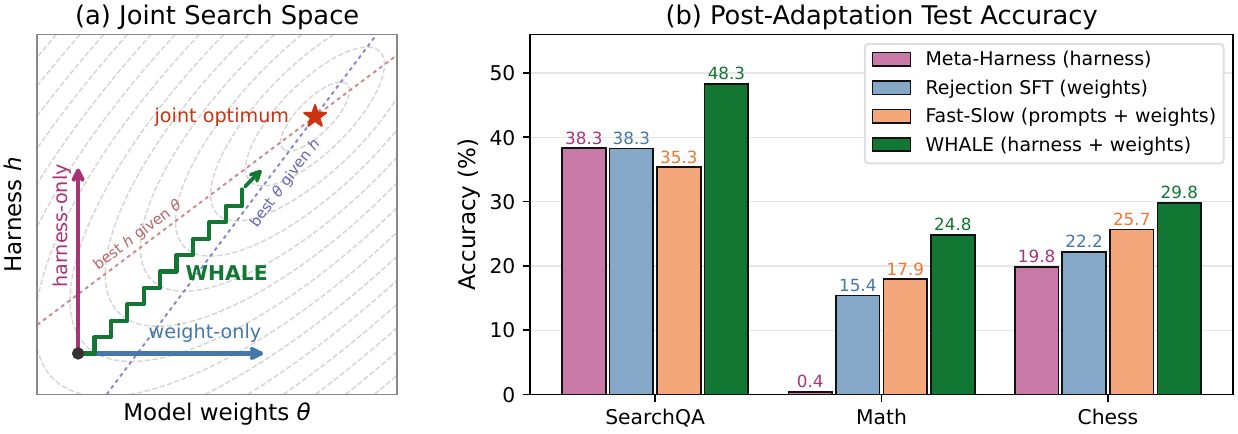}
\caption{Overview of \method{}.
\textbf{(a)} Weight-only and harness-only adaptation move along a single axis of $J(\theta,h)$, whereas alternating updates allow both components to co-adapt.
\textbf{(b)} Our simple joint harness--weight optimization recipe outperforms single-component updates across three diverse domains, as well as joint prompt--weight optimization.
}
\label{fig:hero}
\end{figure*}

\section{Introduction}
\label{sec:introduction}

Model weights are only half of an agentic language-model system. The other half is the harness, the code that decides which observations enter the context, how tools are exposed and invoked, how execution errors are handled, and when an interaction terminates. 
Consider a question-answering agent with a retrieval tool: stronger weights cannot use evidence that a brittle harness never retrieves, while better retrieval cannot help a model that cannot synthesize the returned evidence.
In this paper, we aim to address this shifting bottleneck by jointly optimizing the harness and the model for the given problem.

Prior work has jointly adapted model weights and their surrounding context, but has restricted the latter to textual prompts~\citep{soylu2024bettertogether,zhang2026espl,tiwari2026fst}. 
Executable tool-using agents expose a broader target for adaptation: the harness code governing tools, observations, error handling, and control flow. 
Recent LLM-based optimizers make direct search over this code feasible~\citep{hu2025automateddesignagenticsystems,lin2026agentic,lee2026metaharness}.
Although both components can now be optimized directly, how to coordinate model training with full-harness search remains unclear.

Coordinating model training with harness search is difficult because the two phases operate on different timescales.
Harness search uses a long proposal-and-accept cycle conditioned on many rollouts, whereas online fine-tuning alternates rollout collection with gradient updates.
Running the phases concurrently either evaluates each update against a moving counterpart, thereby confounding credit assignment, or requires synchronization barriers due to their mismatched update cadences.
Alternation avoids this tradeoff by decomposing the coupled problem into conditional updates, each performed with respect to a fixed counterpart~\citep{wright2015coordinate}. 
This raises a key scheduling question: how much should each component change before switching? Each phase must gather enough evidence to distinguish genuine improvements from noise, but stop before over-optimizing against a fixed counterpart.

To study the design decisions surrounding alternating weight--harness updates, we propose \textbf{W}eight-\textbf{H}arness \textbf{A}lternating \textbf{LE}arning (\textbf{\method{}}), a modular framework that alternates model training under the current harness with harness search under the updated model (Figure~\ref{fig:hero}(a)).
In our experiments, we instantiate the weight-update and harness-search phases with online rejection-sampling fine-tuning (RSFT)~\citep{dong2023raft,gulcehre2023rest,shao2024deepseekmath} and Meta-Harness~\citep{lee2026metaharness}, respectively.
By default, fixed hyperparameters shared across cycles determine the duration of each phase. 
We also propose adaptive \method{}, which allows phase durations to vary across cycles by switching, after a minimum duration, when the current phase's training signal stops improving.

We evaluate \method{} against weight-only, harness-only, and Fast--Slow Training (FST, \citet{tiwari2026fst}) across three domains: search question answering (SearchQA), mathematical reasoning with code execution (Math), and chess puzzles.
To isolate the effect of the harness search space, our FST implementation uses the same update methods and schedule as \method{} but restricts the search to the system and user prompts.
\method{} improves over the single-component baselines by 7.67--24.38 percentage points and over FST by 4.15--13.00 points (Figure~\ref{fig:hero}(b)).
Across domains, we observe both harness-limited and weight-limited regimes: in SearchQA, harness search alone matches peak weight-only accuracy with far fewer rollouts, whereas in Math it yields almost no improvement until a small weight update makes the same harness search effective. 
These contrasting regimes show why joint optimization matters: the bottleneck depends on the domain, and updating one component can unlock gains from the other.
Compared with stagewise optimization, which spends the full weight-update budget before running the full harness-search budget once, \method{}'s small alternating updates improve accuracy by 5.32 and 9.16 percentage points in SearchQA and Math, respectively, and surpass the stagewise result after only 29\% and 49\% as many rollouts. 
We also find that adaptive \method{} outperforms the main fixed schedule (with tuned hyperparameters) in both domains we evaluate.
Together, our findings provide a controlled, budget-matched study of alternating weight-harness updates and motivate future work on joint weight-harness optimization.

\section{Related Work}
\label{sec:related}

\textbf{Training for multi-turn reasoning and tool use.}
Language-model agents commonly interleave reasoning with actions~\citep{yao2023react}, and recent methods train this behavior in search and code-execution environments~\citep{nakano2022webgptbrowserassistedquestionansweringhuman,chen2023fireactlanguageagentfinetuning,jin2025searchr1,feng2025retool}. Prior work on reward-ranked and reinforced self-training methods motivate SFT-style, reward-filtered updates as simpler and more stable alternatives to policy gradient methods~\citep{zelikman2022starbootstrappingreasoningreasoning,singh2024humandatascalingselftraining,dong2023raft,gulcehre2023rest,NEURIPS2023_23e6f78b}. We use rejection-sampling fine-tuning from this family for the weight-update phase. Existing methods for training reasoning agents either keep the harness fixed or abstract it away, whereas we study how to train model weights as the harness co-adapts.

\textbf{Optimizing systems around fixed models.}
Work outside model weights optimizes increasingly expressive artifacts, from natural-language instructions and modular prompt programs~\citep{yang2024opro,khattab2024dspy,agrawal2025gepa} to agent architectures and workflows~\citep{hu2024adas,zhang2024aflow,zhuge2024language,shang2025agentsquareautomaticllmagent}.
Recent methods extend this search to executable harnesses and their components through source-code editing, configuration search, and trajectory-driven updates~\citep{lin2026agentic,sengupta2026harborautomatedharnessoptimization,pan2026evolvingagentsdarkretrospective,park2026autosaddlerautomaticharnessoptimization}.
We use Meta-Harness~\citep{lee2026metaharness} as a simple, modular harness-search method that composes directly with weight updates.
\method{} connects these lines of work by alternating reward-filtered model training with executable harness search.

% Prompt optimizers such as OPRO~\citep{yang2024opro} and program optimizers such as DSPy~\citep{khattab2024dspy} improve natural-language instructions or modular programs within a fixed, developer-written pipeline structure. GEPA evolves prompt components from execution traces and natural-language reflection while keeping model weights fixed~\citep{agrawal2025gepa}. Automated Design of Agentic Systems and AFlow search over richer agent architectures and workflows~\citep{hu2024adas,zhang2024aflow}. Meta-Harness extends this direction to executable agent code: a proposer inspects source code, evaluation scores, and interaction traces, then edits the harness in an outer optimization loop~\citep{lee2026metaharness}; we use it as our harness-search operator.

\textbf{Joint and alternating optimization.}
Alternating optimization methods address coupled problems by updating one component while holding the others fixed~\citep {wright2015coordinate}. Weight and harness adaptation has this structure: each component shapes the trajectories used to improve the other, so fixing one prevents its drift from confounding the active update.
Prior methods that jointly adapt prompts and model weights differ in how they coordinate the two updates, but these works restrict the context-side variable to text~\citep{soylu2024bettertogether,ziems2025composing,bo2026metatuner,lu2026p2o,zhang2026espl,tiwari2026fst}.
\method{} instead searches over executable harness programs, motivated by recent work showing that non-prompt harness choices, including tool interfaces, orchestration, memory, and middleware, can substantially alter fixed-model behavior~\citep{yang2024swe,zhuge2024language,lee2026metaharness,lin2026agentic}.
Our FST control~\citep{tiwari2026fst} uses the same update methods and schedule as \method{} while restricting harness search to the system and user prompts, isolating this difference in expressivity.
% BetterTogether composes prompt optimization with fine-tuning, including by alternating the two, for modular language-model pipelines~\citep{soylu2024bettertogether}, and a follow-up extends the composition to online policy-gradient training~\citep{ziems2025composing}. E-SPL interleaves reinforcement-learning weight updates with an evolving population of system prompts~\citep{zhang2026espl}, and FST treats the prompt as fast weights alongside slow model parameters~\citep{tiwari2026fst}, which we include as a baseline (Section~\ref{sec:experimental-settings}). Our study extends this line from textual prompts to executable harness programs and ablates the alternation schedule itself.

\section{Preliminaries}
\label{sec:preliminaries}

Let $\theta\in\Theta$ be the model parameters and $h\in\mathcal{H}$ an executable harness program. The harness includes the system instructions, tool schemas, context management, parser and execution logic, and termination policy. Together, $\theta$ and $h$ induce a conditional distribution $\pi_{\theta,h}(\cdot\mid x)$ over trajectories $\tau \in \mathcal{T}$
comprising model messages, tool calls and results, and environment transitions. Given a task distribution $\mathcal{P}$ and verifier $R: \mathcal{T}\to[0,1]$, we can define the expected reward of the system as:
\begin{equation}
    J(\theta,h) \triangleq
    \mathbb{E}_{x\sim\mathcal{P}}
    \mathbb{E}_{\tau\sim\pi_{\theta,h}(\cdot\mid x)}
    [R(x,\tau)].
    \label{eq:joint-objective}
\end{equation}
We seek a model--harness pair $(\theta^\star,h^\star)$ that maximizes Equation~\ref{eq:joint-objective}.
Rather than optimizing this objective directly over $\Theta\times\mathcal{H}$, we separate adaptation into a weight-update phase that updates $\theta$ with $h$ fixed and a harness-search phase that updates $h$ with $\theta$ fixed.
The two remain coupled through $\pi_{\theta,h}$, since each update changes the policy on which the other operates.
Many algorithms can instantiate either update; in this work, we use online RSFT and Meta-Harness, respectively, as described below.

\subsection{Weight Updates: Online Rejection-Sampling Fine-Tuning}
\label{sec:rsft}

We use online rejection-sampling fine-tuning (RSFT) for weight updates: an online variant of rejection-sampling fine-tuning~\citep{shao2024deepseekmath} that trains the model with supervised learning only on its own verifier-accepted rollouts. Given a fixed harness $h$, let $\theta$ denote the parameters being updated and $\theta_{\mathrm{old}}$ the frozen copy that generates the rollouts at rollout step $s$. For a batch $\mathcal{B}_s\subseteq\mathcal{D}_{\mathrm{weight}}$, RSFT samples $G$ trajectories per prompt and keeps only the verifier-accepted pairs: $\mathcal{S}_{s}^{+}$ collects every $(x,\tau)$ with $x\in\mathcal{B}_{s}$, $\tau\sim\pi_{\theta_{\mathrm{old}},h}(\cdot\mid x)$, and $R(x,\tau)=1$. For a trajectory $\tau$, let $\mathcal{I}(\tau)$ denote the positions of model-generated tokens, excluding user prompts and tool results. RSFT maximizes the token-normalized supervised log-likelihood
\begin{equation}
    \widehat{J}_{\mathrm{weight}}(\theta;\theta_{\mathrm{old}},h)
    =
    \frac{1}{
        \displaystyle
        \sum_{(x,\tau)\in\mathcal{S}_{s}^{+}}
        |\mathcal{I}(\tau)|
    }
    \sum_{(x,\tau)\in\mathcal{S}_{s}^{+}}
    \sum_{t\in\mathcal{I}(\tau)}
    \log p_{\theta}\left(\tau_t\mid x,h,\tau_{<t}\right).
    \label{eq:rsft}
\end{equation}
In implementation, RSFT ascends $\widehat{J}_{\mathrm{weight}}$ with minibatch stochastic gradient steps: at rollout step $s$, it draws SFT minibatches $\mathcal{M}\subseteq\mathcal{S}_s^+$, applies gradient updates, and synchronizes the updated weights to the rollout workers before collecting the next group.

\subsection{Harness Search: Meta-Harness}
\label{sec:mh}

We use Meta-Harness (MH)~\citep{lee2026metaharness} for harness search: an iterative proposal--evaluation--selection search over executable harness programs. Given fixed model parameters $\theta$, MH searches for a better harness over $I$ iterations, starting from an initial harness $h^{\mathrm{init}}\in\mathcal{H}$. Let $\mathcal{A}_j$ denote the accumulated harness archive at iteration $j$, seeded with $h^{\mathrm{init}}$. Evaluating a harness $h$ yields an artifact $e_h$, comprising its aggregate score, per-example outcomes from the fixed binary verifier $R$, and saved trajectory logs; $\mathcal{E}_j=\{e_h : h\in\mathcal{A}_j\}$ collects the artifacts of the archived harnesses. The proposer selectively inspects these artifacts and the archived source files through filesystem tools, and one proposer session writes $M$ candidate harnesses $\mathcal{C}_j=\operatorname{Propose}(\mathcal{A}_j,\mathcal{E}_j;M)\subseteq\mathcal{H}$; the archive and its artifacts then grow to $\mathcal{A}_{j+1}=\mathcal{A}_j\cup\mathcal{C}_j$ and $\mathcal{E}_{j+1}=\mathcal{E}_j\cup\{e_h : h\in\mathcal{C}_j\}$. Each candidate $h\in\mathcal{C}_j$ is evaluated on $\mathcal{D}_{\mathrm{harness}}$ with the rollout-based empirical score
\begin{equation}
    \widehat{J}_{\mathrm{harness}}(\theta,h)
    =
    \frac{1}{|\mathcal{D}_{\mathrm{harness}}|}
    \sum_{x\in\mathcal{D}_{\mathrm{harness}}}
    \widehat{\mathbb{E}}_{\tau\sim\pi_{\theta,h}(\cdot\mid x)}
    [R(x,\tau)].
    \label{eq:mh-score}
\end{equation}
Here, $\widehat{\mathbb{E}}$ denotes the empirical mean over the trajectories sampled for each example; our experiments use one trajectory per example during harness search.
MH then selects the accepted harness: $h^{\mathrm{accept}} \in \arg\max \ \widehat{J}_{\mathrm{harness}}(\theta,h)$, where the maximization is over the archive $\mathcal{A}_{j+1}$.
After each iteration, MH marks the highest-scoring harness in the expanded archive as accepted. Subsequent proposals can inspect the full archive and its artifacts, allowing them to refine any prior candidate or revert to an earlier design. After $I$ iterations, MH returns the highest-scoring archived harness.

\section{\method{}: Weight-Harness Alternating Learning}
\label{sec:method}

Weight-Harness Alternating Learning (\method{}) consists of two alternating phases, each improving one component while holding the other fixed: each cycle trains the model weights under the current harness, then searches for a better harness under the updated model, and repeats. \method{} requires only black-box interfaces for the two update procedures, and can therefore accommodate any weight-update and harness-search method. Starting from an initial model--harness pair $(\theta_0,h_0)$, cycle $k$ computes
\begin{equation}
    \theta_{k+1} = \operatorname{ModelUpdate}(\theta_k; h_k,\mathcal{D}_{\mathrm{weight}}),
    \qquad
    h_{k+1} = \operatorname{HarnessSearch}(h_k; \theta_{k+1},\mathcal{D}_{\mathrm{harness}}).
    \label{eq:alternation}
\end{equation}
Budgets and other algorithm-specific settings belong to the instantiation. We instantiate $\operatorname{ModelUpdate}$ with RSFT (Section~\ref{sec:rsft}) and $\operatorname{HarnessSearch}$ with MH (Section~\ref{sec:mh}). For RSFT, the weight-update phase budget $E$ is the number of training epochs: cycle $k$ consumes the span $\mathcal{D}_{\mathrm{weight}}^{[kE:(k+1)E)}$ from a persistent sampler. For MH, the harness-search phase budget $I$ is the number of harness-search iterations, and $M$ is the number of candidate harnesses proposed per iteration. Algorithm~\ref{alg:alternating-rsft-mh}\atapp{app:algorithms} details the \method{} procedure for our instantiation with RSFT and MH.

\ifarxiv\begin{algorithm}[!ht]
\caption{\method{} with RSFT and MH}
\label{alg:alternating-rsft-mh}
\small
\begin{algorithmic}[1]
\REQUIRE $(\theta_0,h_0)$; datasets $\mathcal{D}_{\mathrm{weight}}$, $\mathcal{D}_{\mathrm{harness}}$; fixed binary verifier $R$; budgets $E$, $I$, $M$
\FOR{$k=0,1,\ldots$}
    \STATE $\theta\leftarrow\theta_k$
    \FOR{each prompt batch $\mathcal{B}_s$ in $\mathcal{D}_{\mathrm{weight}}^{[kE:(k+1)E)}$}
        \STATE $\theta_{\mathrm{old}}\leftarrow\theta$
        \STATE collect $\mathcal{S}_s^+$ under $(\theta_{\mathrm{old}},h_k)$
        \STATE ascend $\widehat{J}_{\mathrm{weight}}(\theta;\theta_{\mathrm{old}},h_k)$ over minibatches of $\mathcal{S}_s^+$ (Eq.~\ref{eq:rsft})
    \ENDFOR
    \STATE $\theta_{k+1}\leftarrow\theta$;\quad $h^{\mathrm{init}}\leftarrow h_k$;\quad $\mathcal{A}_0\leftarrow\{h^{\mathrm{init}}\}$;\quad $\mathcal{E}_0\leftarrow\{e_{h^{\mathrm{init}}}\}$
    \FOR{$j=0,\ldots,I-1$}
        \STATE $\mathcal{C}_j\leftarrow\operatorname{Propose}(\mathcal{A}_j,\mathcal{E}_j;M)$
        \STATE evaluate $\widehat{J}_{\mathrm{harness}}(\theta_{k+1},h)$ for each $h\in\mathcal{C}_j$ (Eq.~\ref{eq:mh-score}), yielding artifacts $e_h$
        \STATE $\mathcal{A}_{j+1}\leftarrow\mathcal{A}_j\cup\mathcal{C}_j$;\quad $\mathcal{E}_{j+1}\leftarrow\mathcal{E}_j\cup\{e_h : h\in\mathcal{C}_j\}$
        \STATE $h^{\mathrm{accept}}\leftarrow\underset{h\in\mathcal{A}_{j+1}}{\arg\max}\ \widehat{J}_{\mathrm{harness}}(\theta_{k+1},h)$
    \ENDFOR
    \STATE $h_{k+1}\leftarrow h^{\mathrm{accept}}$
\ENDFOR
\RETURN the pair $(\theta_k,h_k)$ with the best accuracy on $\mathcal{D}_{\mathrm{test}}$
\end{algorithmic}
\end{algorithm}
\fi

The two phases use different update data and local objectives, but are coupled through the induced trajectory distribution $\pi_{\theta,h}$: the weight-update phase adapts the model to the prompts, tools, observations, and termination policy defined by $h_k$, changing behavior such as response length and reasoning patterns, and the harness-search phase then searches for a harness better matched to the updated $\theta_{k+1}$. Repeating the two lets the components co-adapt.

\paragraph{Adaptive \method{}}
replaces these fixed phase budgets with a per-phase stopping rule. We monitor the training metrics of each phase. After a minimum phase length, the weight-update phase stops once its training reward, averaged over a sliding window of recent steps, has not improved for a fixed number of steps. The harness-search phase stops once the best training score in its archive has not improved for a fixed number of iterations. The rule is early stopping in spirit, but it is applied per phase rather than per run and is driven only by training signals, never by a validation dataset. This removes the schedule $(E,I)$ from the hyperparameters. Algorithm~\ref{alg:adaptive-whale} in Appendix~\ref{app:adaptive-algorithm} details the procedure and its stopping parameters, and Section~\ref{sec:adaptive-schedule} evaluates this adaptive schedule.

\section{Experiments}
\label{sec:experiments}

Our experiments ask whether alternating weight updates with full-harness search outperforms optimizing either component alone, and whether full-harness search offers gains beyond prompt-only adaptation.
We test these questions across three tool-using domains, using shared initializations and matched update budgets within each domain to isolate the effect of joint adaptation.

\subsection{Domains and Data}
\label{sec:experimental-settings}

\paragraph{SearchQA.}
Following Search-R1~\citep{jin2025searchr1}, the weight-update training dataset $\mathcal{D}_{\mathrm{weight}}$ contains 18,946 questions: 14,801 from HotpotQA~\citep{yang2018hotpotqa} and 4,145 from Natural Questions~\citep{kwiatkowski2019naturalquestions}. The harness-search training dataset $\mathcal{D}_{\mathrm{harness}}$ comprises 256 questions, 200 from HotpotQA and 56 from Natural Questions, matching the $\mathcal{D}_{\mathrm{weight}}$ distribution. The test dataset $\mathcal{D}_{\mathrm{test}}$ contains 700 questions, with 100 each from 2WikiMultiHopQA~\citep{ho2020twowiki}, Bamboogle~\citep{press2023compositionality}, HotpotQA, MuSiQue~\citep{trivedi2022musique}, Natural Questions, PopQA~\citep{mallen2023popqa}, and TriviaQA~\citep{joshi2017triviaqa}. Natural Questions, PopQA, and TriviaQA are single-hop, whereas HotpotQA, 2WikiMultiHopQA, MuSiQue, and Bamboogle are multi-hop~\citep{jin2025searchr1}. The retrieval environment answers an agent's query with the top-$k$ nearest documents from the Wikipedia 2018 corpus~\citep{karpukhin2020dpr}, indexed using FAISS~\citep{johnson2017faiss}.

\paragraph{Mathematical Reasoning.}
The weight-update training dataset $\mathcal{D}_{\mathrm{weight}}$ contains 17,917 problems from DAPO-Math-17K~\citep{yu2025dapo}. The harness-search training dataset $\mathcal{D}_{\mathrm{harness}}$ contains 256 problems randomly selected from DAPO-Math-17K. The test dataset $\mathcal{D}_{\mathrm{test}}$ comprises AIME 2024~\citep{maxwelljia2024aime} and AIME 2025~\citep{yentinglin2025aime} problems. Following ReTool~\citep{feng2025retool}, the agent may write Python programs for intermediate computation; a CPU environment executes them and returns their output or error text as the tool response.

\paragraph{Chess Puzzles.}
All three datasets are derived from the Lichess open puzzle database~\citep{lichess2026database}. $\mathcal{D}_{\mathrm{weight}}$, $\mathcal{D}_{\mathrm{harness}}$, and $\mathcal{D}_{\mathrm{test}}$ hold 16,384, 256, and another 256 puzzles, mutually disjoint. The agent generates chess moves in Universal Chess Interface (UCI) notation and passes them to an environment that maintains the board state. An incorrect legal move terminates the rollout immediately, whereas an illegal move terminates it after the permitted retry is exhausted. After a correct nonterminal move, the environment returns the puzzle's predefined fixed opponent reply as the tool response. A correct terminal move completes the reference solution sequence and ends the rollout.

Appendix~\ref{app:implementation} specifies each domain's harness-search space, initial harness $h_0$ and binary verifier $R$. In every domain, the harness search may modify the system and user prompts, the formatting of tool inputs and outputs, the feedback returned after a tool call, and the stopping criteria and turn allocation, while the model, datasets, verifier, environment transition rules, and reference answers stay fixed. The verifier $R(x,\tau)\in\{0,1\}$ is a sparse trajectory-level reward: intermediate steps receive no reward, and $R(x,\tau)=1$ only when the terminal task outcome is correct.

\subsection{Evaluation Setup}

SearchQA and Math use Qwen3.5-2B; Chess Puzzles use Qwen3.5-4B~\citep{qwen2026qwen35}. Every method starts from the same domain-specific $(\theta_0,h_0)$, specified in Appendix~\ref{app:harness-space}. Weight-only holds $h=h_0$ and trains for 4, 6, and 4 epochs in SearchQA, Math, and Chess Puzzles; harness-only holds $\theta=\theta_0$ and runs 40, 60, and 40 harness-search iterations; \method{} alternates the two under $(E,I)=(0.6,6)$. 
All rollouts use a sampling temperature of 1.0, top-$p$ of 1.0, and top-$k$ of 20. The response lengths are 8,192, 8,192, and 16,384 tokens for SearchQA, Math, and Chess Puzzles, respectively. The weight update samples $G=8$ trajectories per prompt, the harness search 1 rollout per candidate--example pair, and test evaluation 8 trajectories per example, reported as \meanatk{}.

Adaptive \method{} keeps this configuration but replaces the fixed budgets with the patience rule of Section~\ref{sec:method}: the weight-update phase uses a minimum phase length, patience, and averaging window of 0.2 epochs each, and the harness-search phase runs at least 6 iterations and stops after 2 iterations without a new best training score.
FST~\citep{tiwari2026fst} adapts prompts alongside weights. We run it as a prompt-restricted control, using the update methods and schedule of \method{} but with harness search confined to the system and user prompts of $h_0$, thereby isolating search expressivity. The single-component baselines receive budgets matching the cumulative budgets of the \method{} and FST runs; Table~\ref{tab:principal-config} in Appendix~\ref{app:implementation} shows details for all configurations.

\subsection{Does Joint Weight--Harness Optimization Improve Performance?}
\label{sec:results}

\begin{figure*}[t]
    \centering
    \includegraphics[width=0.92\textwidth]{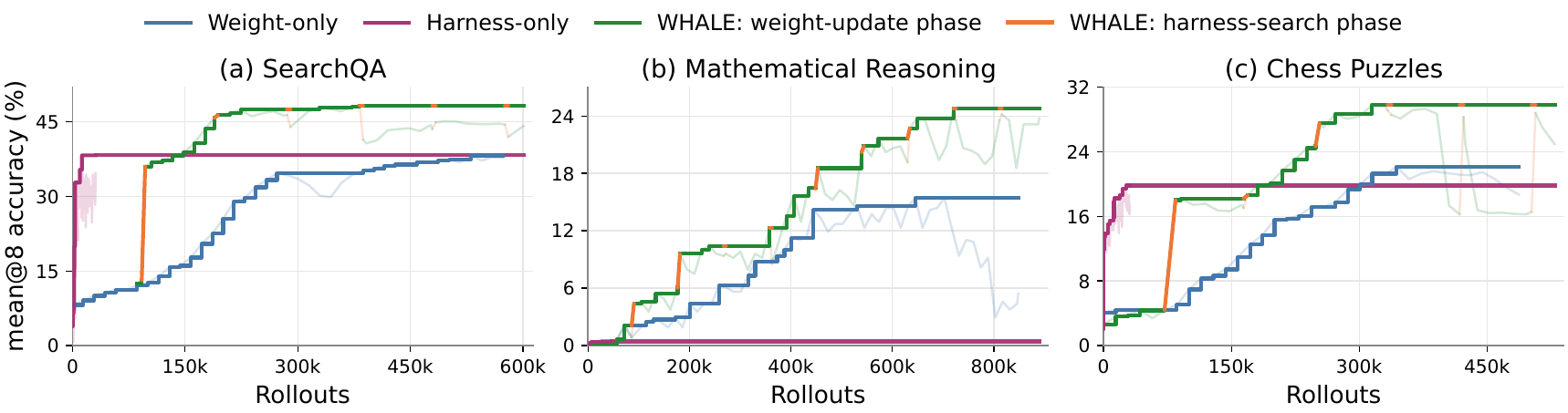}
    \caption{Best-so-far test \meanatk{} accuracy on the test datasets as rollouts accumulate. Harness-only is held constant after 40, 60, and 40 iterations in SearchQA, Mathematical Reasoning, and Chess Puzzles, respectively.}
    \label{fig:main-results}
\end{figure*}

Figure~\ref{fig:main-results} shows test \meanatk{} accuracy for single-component baselines and WHALE.
Using one fixed schedule, $(E,I)=(0.6,6)$, \method{} achieves the highest accuracy in all three domains.
The ordering of the single-component baselines changes across domains: weight-only and harness-only are effectively tied in SearchQA, whereas weight-only is substantially stronger in Math and Chess Puzzles.
\method{} improves under both orderings, outperforming the stronger single-component baseline by 7.67--10.05 percentage points.
FST exhibits a different pattern: prompt--weight adaptation underperforms both single-component baselines in SearchQA but outperforms them in Math and Chess Puzzles.
\method{} nevertheless exceeds FST by 4.15--13.00 points (Figure~\ref{fig:hero}(b)).
Because our FST control uses the same update methods and schedule while restricting harness search to prompts, this comparison measures the benefit of adapting the broader executable harness.
The gains over the single-component baselines also hold on every individual benchmark; Table~\ref{tab:main-results}\atapp{app:per-benchmark} reports the full per-dataset results.

\begin{findingbox}
\textbf{Consistent gains across domains.} 
With one fixed schedule, \method{} outperforms weight-only, harness-only, and prompt-restricted FST across all three domains, despite differences in which component is strongest alone.
\end{findingbox}

\section{Analysis}
\label{sec:analysis}

The aggregate gains mask different interactions between model weights and harness code across domains.
We first identify which component limits performance and how updating one changes the gains available to the other.
We then test whether small interleaved updates improve over stagewise optimization, how phase duration affects performance, and whether training signals can determine when to switch.

\subsection{Which Component Limits Performance?}
\label{sec:task-dependent-roles}
\begin{figure*}[t]
    \centering
    \includegraphics[width=\textwidth]{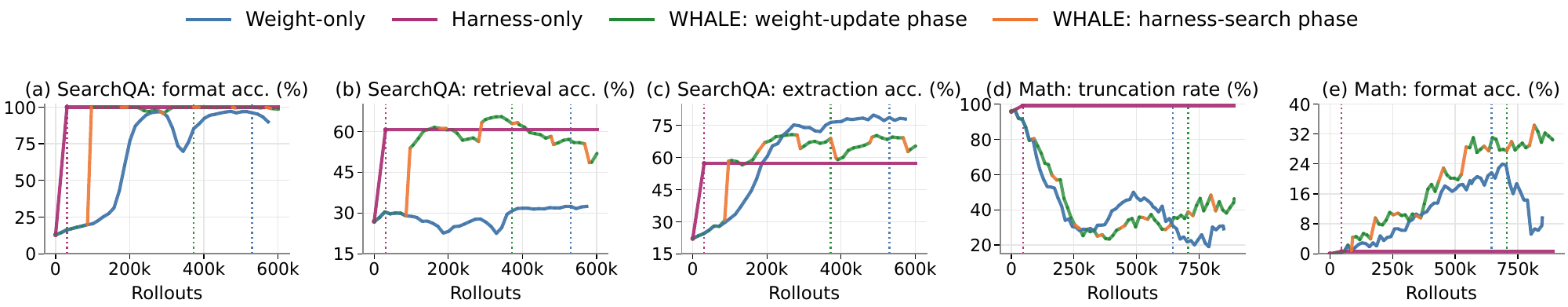}
    \caption{Behavior metrics on test trajectories as rollouts accumulate. \textbf{(a--c)} SearchQA: the fraction of responses following the required answer format, the fraction of trajectories retrieving at least one document containing the reference answer, and correctness conditional on retrieving such a document. \textbf{(d--e)} Mathematical Reasoning: the fraction of responses reaching the token limit, and the fraction from which the verifier extracts a final boxed answer. Dotted vertical lines mark where each method achieves its best test \meanatk{} accuracy.}
    \label{fig:behavior}
\end{figure*}

We observe two domain-dependent bottleneck regimes.
SearchQA is \emph{harness-dominant}: harness search matches the peak accuracy of weight updates using only 5.79\% as many rollouts.
Math is \emph{model-dominant}: harness search reaches only 0.42\%, compared with 15.42\% for weight updates.
These comparisons count target-agent rollouts and exclude proposer compute in the harness search phase.

\textbf{SearchQA is harness-dominant}.
Figure~\ref{fig:behavior}(a--c) quantifies three capabilities required for SearchQA. Both weight updates and harness search improve format compliance, with weight-only and harness-only peaking at 97.07\% and 99.98\%: weight updates learn the behavior from format-correct on-policy trajectories, whereas harness search emphasizes the answer format in the user prompt and requires an answer rather than another search on the final turn. Harness-only reaches the higher value with fewer rollouts, so harness search is the more efficient channel here.

For retrieval accuracy, weight-only improves it only modestly, whereas harness-only raises it from 26.88\% to 60.61\% with far fewer rollouts: the model writes the query, but the harness post-processes it and controls which documents return and how many. Combining both, \method{} reaches the highest retrieval accuracy of 65.41\%.
Answer extraction reverses the balance, with weight-only and harness-only peaking at 79.93\% and 57.28\%. In \method{}, the cycle 1 harness-search phase improves extraction sharply, and the cycle 2 weight-update phase improves it further, but from cycle 3 onward the harness-search phases hold it back.

\textbf{Mathematical reasoning is weight-dominant.}
Figure~\ref{fig:behavior}(d--e) quantifies two capabilities required for Math, where the bottleneck is response truncation. Weight-only cuts the truncation rate from 95.83\% to 30.83\%, learning to shorten responses from correct on-policy trajectories that terminate before the token limit. Harness-only attempts the same through response caps, turn limits, and a final-answer recovery step, but cannot overcome the base model's behavior, which is also why its format accuracy barely moves. \method{} does shorten responses, reducing truncation and improving format accuracy, and its cycle 1 harness-search phase is the clearest case: in 4,608 rollouts it raises format accuracy from 0.83\% to 4.38\%, a gain of 3.54 percentage points, where harness-only spends 46,080 rollouts for a gain of 0.63 points (0.00\% to 0.63\%).

\begin{findingbox}
\textbf{Dominance regimes.} 
We observe both \emph{harness-dominant} and \emph{model-dominant} regimes, and for capabilities that both weight updates and harness search can improve, harness search is more rollout-efficient. A short harness search therefore serves as a low-cost diagnostic before extensive model training. Conversely, a small weight update can make harness search effective where it previously was not.
\end{findingbox}

% \begin{findingbox}
% \textbf{Finding 3 (Small weight updates unlock harness search).} A small weight update can unlock an otherwise ineffective harness search: in model-dominant Math, harness search alone yields almost no improvement, yet after a small weight update the same harness search becomes highly effective.
% \end{findingbox}

\FloatBarrier

\subsection{How Should Weight and Harness Updates Be Scheduled?}
\label{sec:why-alternate}

We ablate the alternation schedule $(E,I)$, the per-cycle weight-update epoch and harness-search iteration budgets, in SearchQA and Math, the harness-dominant and model-dominant extremes. Table~\ref{tab:schedule-best} in Appendix~\ref{app:schedule-best} lists the best point of every run.

\subsubsection{Alternating versus Stagewise Optimization}
\label{sec:stagewise}

The common way to combine the two is stagewise, with one pass per component: train under $h_0$ with the full weight-only budget, then run harness search under the resulting model for the full harness-only budget. Each phase selects its best point based on the training score; at the budgets, we denote $(E^{\ast},I^{\ast})$.

\begin{figure*}[t]
\centering
\includegraphics[width=0.92\textwidth]{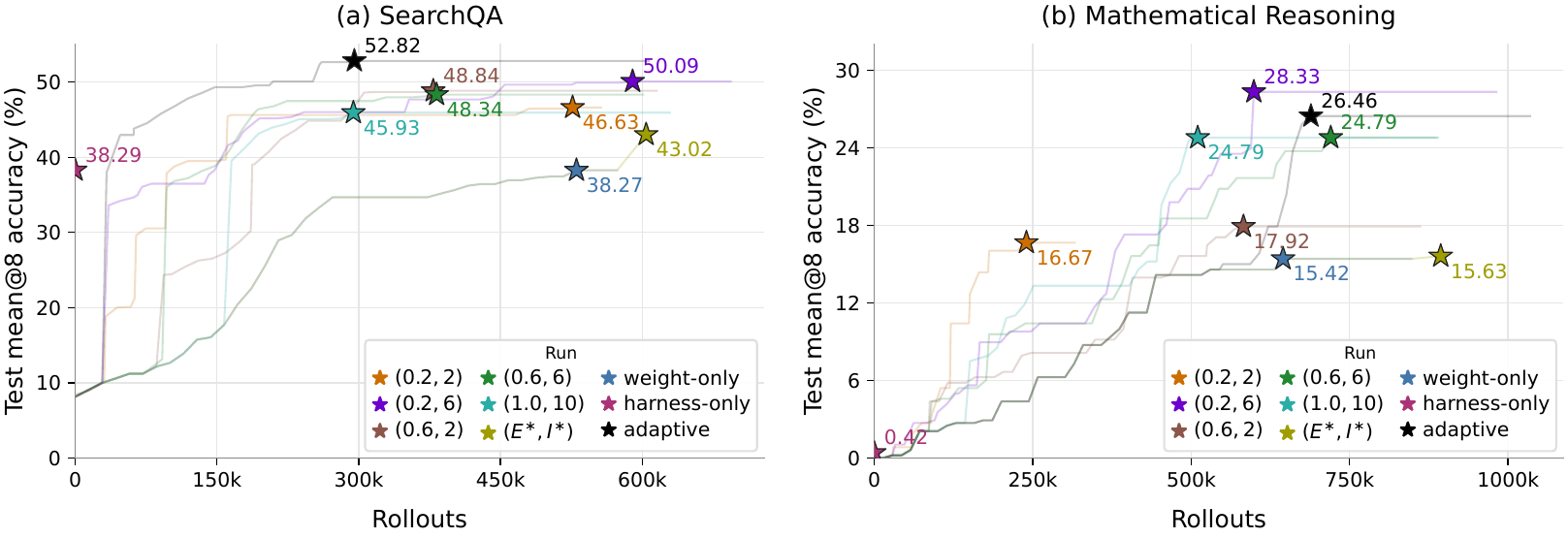}
\caption{Schedule comparison for (a) SearchQA and (b) Mathematical Reasoning. Each run---the five fixed $(E,I)$ schedules, stagewise $(E^{\ast},I^{\ast})$, the weight-only and harness-only baselines, and adaptive \method{}---is marked by a star at its best test \meanatk{} accuracy; faint lines trace each run's best-so-far accuracy.}
\label{fig:cadence-ablation}
\end{figure*}

In SearchQA, $(E^{\ast},I^{\ast})=(2.7,32)$ within the 4-epoch and 40-iteration budgets, reaching 43.02\% test \meanatk{} accuracy (Figure~\ref{fig:cadence-ablation}(a)); in Math, $(E^{\ast},I^{\ast})=(4.3,33)$ within the 6-epoch and 60-iteration budgets, reaching 15.63\% (Figure~\ref{fig:cadence-ablation}(b)). These fall 5.32 and 9.16 points short of \method{} with $(0.6,6)$, which passes the final stagewise accuracy after only 29\% of the stagewise rollouts in SearchQA and 49\% in Math.

We attribute the weaker stagewise performance to conditional over-optimization (Figure~\ref{fig:stagewise-concept}). Because $J(\theta,h)$ (Eq.~\ref{eq:joint-objective}) is not separable, each phase optimizes against a counterpart that later changes. A long weight-update phase over-optimizes the model's behavior for $h_0$, leaving it poorly adapted to the harness produced by the subsequent harness search. A long harness-search phase can likewise over-optimize a harness to $\mathcal{D}_{\mathrm{harness}}$: in Math, the 60-iteration stagewise harness search keeps improving its training score, yet gains only +0.21 percentage points over weight-only on the test set (15.63\% versus 15.42\%).

\begin{findingbox}
\textbf{Alternation dominates stagewise optimization.} 
One large pass per component is dominated by short alternating phases, in both final accuracy and rollout cost, because spending a component's full budget in a single uninterrupted phase over-optimizes it against a counterpart that then changes.
\end{findingbox}

\subsubsection{How Long Should Each Phase Run?}
\label{sec:cadence-default}

Given that alternation should take small steps, we next ask how many updates each cycle should contain. In SearchQA and Math, we compare five schedules $(E,I)\in\{(0.2,2),(0.2,6),(0.6,2),(0.6,6),(1.0,10)\}$ (Figure~\ref{fig:cadence-ablation}). Although the two domains lie at opposite extremes of the regime, the same schedule is strongest in both: $(0.2,6)$ reaches 50.09\% in SearchQA and 28.33\% in Math, exceeding the $(0.6,6)$ schedule in the main comparison by +1.75 and +3.54 percentage points, respectively. An effective schedule must therefore lie between two failure extremes: each phase must gather enough evidence to avoid adopting a noisy choice, yet stop before it over-optimizes against its frozen counterpart. The two ends of the schedule spectrum exhibit exactly these failures, sketched in Figure~\ref{fig:stagewise-concept}.

\noindent
\begin{minipage}[t]{0.40\textwidth}
\vspace{0pt}
\centering
\includegraphics[width=\linewidth]{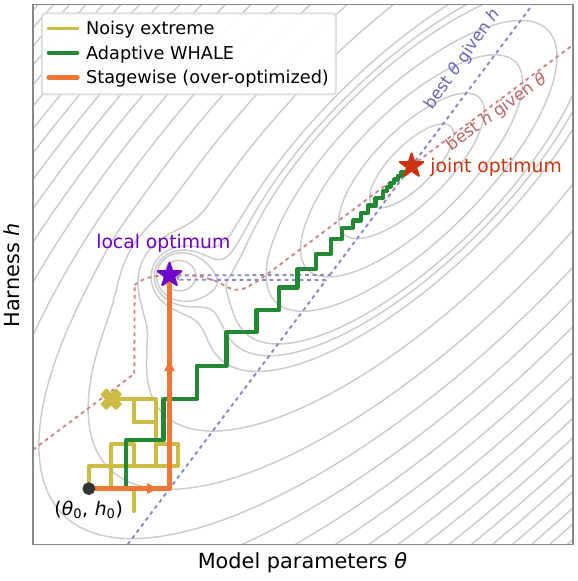}
\captionof{figure}{Running each phase to the maximum of its own line ends at a local optimum (the over-optimized extreme); too little evidence per phase stalls on noise (the noisy extreme); adaptive \method{} tracks the valley to the joint optimum.}
\label{fig:stagewise-concept}
\end{minipage}\hfill
\begin{minipage}[t]{0.57\textwidth}
\vspace{0pt}
\setlength{\parskip}{.5pc}
The noisy extreme is the Math $(0.2,2)$ run. With only two proposal iterations of evidence per cycle, a chance-inflated candidate is accepted, and the model then adapts to it; the run destabilized and was stopped. At the other end, the over-optimization of Section~\ref{sec:stagewise} sets in well before the stagewise limit: scaling the per-cycle budgets up from $(0.2,6)$ never helps in either domain, and along $(0.2,6)$, $(0.6,6)$, $(1.0,10)$ accuracy falls from 50.09\% to 48.34\% to 45.93\% in SearchQA and from 28.33\% to 24.79\% to 24.79\% in Math.

Neither budget can be chosen on its own, because the effect of one depends on the value of the other: in Math, increasing $E$ from 0.2 to 0.6 helps at $I=2$ (16.67\% to 17.92\%) but hurts at $I=6$ (28.33\% to 24.79\%), and the same crossover holds in the $I$ direction in SearchQA. Even in the model-dominant domain, the schedule that runs harness search most often per training epoch prevails, as explained by catalytic interaction (Section~\ref{sec:task-dependent-roles}): each small weight update renews the gains attainable by a bounded harness search.
\end{minipage}

\begin{findingbox}
\textbf{Effective schedules balance noisy selection and over-optimization.} 
Across both domains, the strongest schedule pairs small weight updates with harness search that runs long enough to identify reliable improvements but stops before over-optimizing.
\end{findingbox}

\subsubsection{Can We Automatically Determine When to Switch?}
\label{sec:adaptive-schedule}

Our sweep over schedule hyperparameters identifies a strong range of phase durations but still requires choosing fixed budgets.
We test whether adaptive \method{} can recover this range without fixing $(E,I)$ in advance.
For SearchQA and Math, we compare against the main $(0.6,6)$ schedule and the strongest fixed schedule in our sweep, $(0.2,6)$.
% The findings above prescribe the interval only qualitatively. Adaptive \method{} (Section~\ref{sec:method}) turns this prescription into a stopping rule that brackets the interval from both sides: a small minimum phase length guards the noisy end, and a training-signal patience stop guards the over-optimized end (Section~\ref{sec:experimental-settings}; Figure~\ref{fig:stagewise-concept}). We evaluate it on SearchQA and Math against two references: the $(0.6,6)$ schedule in the main comparison and the best hand-tuned schedule $(0.2,6)$.
As Figure~\ref{fig:cadence-ablation} shows, adaptive \method{} attains the best point of the entire comparison in SearchQA: 52.82\% test \meanatk{} accuracy, +4.48 points over the $(0.6,6)$ schedule and +2.73 points over the best hand-tuned schedule $(50.09\%)$, reached with 23\% fewer rollouts than the $(0.6,6)$ run needs for its own best. In Math, adaptive \method{} reaches 26.46\%: +1.67 points over the $(0.6,6)$ schedule with 4\% fewer rollouts, but 1.87 points below the best hand-tuned schedule (28.33\%).

The realized schedules fall within the region identified by the ablation (Figure~\ref{fig:adaptive-schedule}\atapp{app:realized-schedules}). Median phase lengths are 0.24 and 0.29 weight-update epochs in the two domains, with $I=7$ harness-search iterations per cycle in both, close to the $(0.2,6)$ optimum, while individual phases extend when their training signal keeps improving, up to 1.16 epochs and $I=13$.

\ifarxiv\begin{figure}[!ht]
\centering
\includegraphics[width=\textwidth]{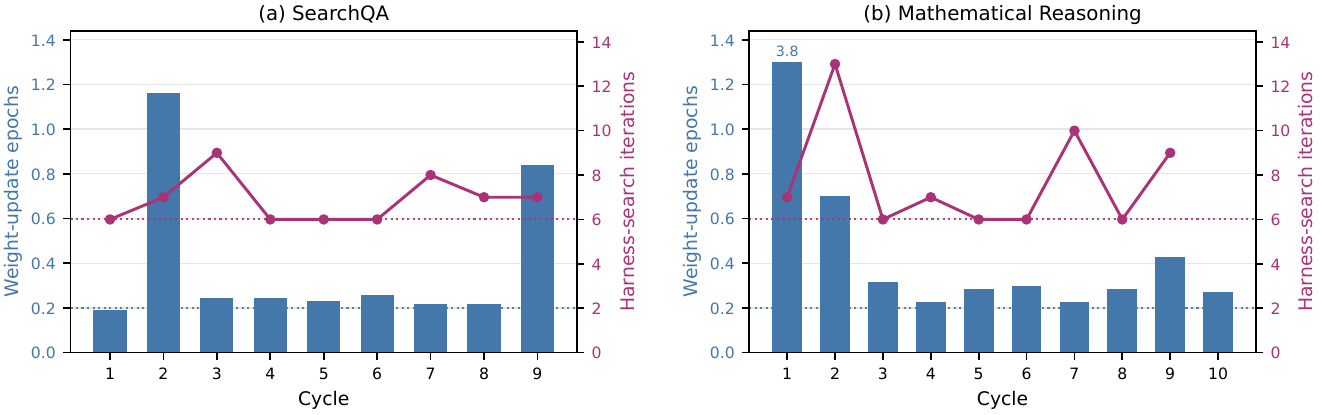}
\caption{Realized per-cycle budgets chosen by the patience rule of adaptive \method{} in (a) SearchQA and (b) Mathematical Reasoning: weight-update epochs (bars, left axis) and harness-search iterations (dots, right axis). Dotted horizontal lines mark the minimum phase lengths (0.2 epoch, $I=6$).}
\label{fig:adaptive-schedule}
\end{figure}
\fi

% \begin{findingbox}
% \textbf{Finding 6 (An automatic schedule from training signals).} The patience rule places every phase between the two failure extremes: the minimum phase length supplies enough evidence to avoid a noisy choice, and the patience stop ends the phase before it over-optimizes. Adaptive \method{} thereby recovers the strongest schedule region without tuning $(E,I)$.
% \end{findingbox}

\section{Conclusion}
\label{sec:conclusion}

We presented \method{}, a modular framework that alternates weight updates with harness search over executable agent harnesses. Across SearchQA, Math, and Chess Puzzles it outperforms the weight-only and harness-only baselines by +7.67 to +24.38 percentage points, and FST, whose harness search is restricted to prompts, by +4.15 to +13.00 points. Our analysis shows that domains divide into harness-dominant and model-dominant regimes, and that a small weight update can unlock an otherwise ineffective harness search. A controlled schedule study further shows that one large pass per component is dominated by small alternating steps, and that per-cycle budgets must lie between the noisy and over-optimized extremes. Finally, a per-phase patience rule on training signals automatically identifies this interval, replacing the fixed budgets.

Future work includes instantiating the two operators with other weight-update and harness-search algorithms, as permitted by the black-box interfaces in Section~\ref{sec:method}, and scaling the study to larger models and additional domains. More broadly, these results motivate treating the model and its harness not as separately engineered artifacts but as one jointly trained system.

% Acknowledgments name people and so would break anonymity; they appear in
% the arXiv preprint only.
\ifarxiv
\subsubsection*{Acknowledgments}

We would like to express our sincere gratitude to Myungseok Oh and Yun Jegal for their valuable contributions and support throughout this work.
\fi

\clearpage
\bibliography{iclr2027_conference}
\bibliographystyle{iclr2027_conference}

\clearpage
\appendix
\section{Algorithms}
\label{app:algorithms}

\subsection{Adaptive \method{}}
\label{app:adaptive-algorithm}

Algorithm~\ref{alg:adaptive-whale} details adaptive \method{} for our instantiation with RSFT and MH, replacing the fixed budgets $E$ and $I$ of Algorithm~\ref{alg:alternating-rsft-mh} with per-phase stopping rules. The weight-update phase uses three parameters, all measured in rollout steps: an averaging window $W$, a minimum phase length $T_{\min}$, and a patience $P_{\mathrm{w}}$. At step $t$ of a phase, the signal $\bar{r}_t$ is the mean verifier reward of the rollouts collected in the most recent $W$ steps, and $\bar{r}^{\,\ast}$ and $t^{\ast}$ track the best signal within the phase and the step that attained it; the phase ends once $t>T_{\min}$ and the signal has not set a new phase best for $P_{\mathrm{w}}$ steps. The harness-search phase uses a minimum number of iterations $J_{\min}$, counting the evaluation of the incoming harness as the first, and a patience $P_{\mathrm{h}}$. Here $J^{\ast}$ tracks the best training score in the archive and $j^{\ast}$ the iteration that attained it; the phase ends once at least $J_{\min}$ iterations have run and $J^{\ast}$ has not improved for $P_{\mathrm{h}}$ iterations. An improvement of any margin counts. Our experiments set $W=T_{\min}=P_{\mathrm{w}}=0.2$ epoch and $(J_{\min},P_{\mathrm{h}})=(6,2)$.

\begin{algorithm}[!ht]
\caption{Adaptive \method{} with RSFT and MH}
\label{alg:adaptive-whale}
\small
\begin{algorithmic}[1]
\REQUIRE $(\theta_0,h_0)$; datasets $\mathcal{D}_{\mathrm{weight}}$, $\mathcal{D}_{\mathrm{harness}}$; fixed binary verifier $R$; budget $M$; stopping parameters $(W,T_{\min},P_{\mathrm{w}})$ and $(J_{\min},P_{\mathrm{h}})$
\FOR{$k=0,1,\ldots$}
    \STATE $\theta\leftarrow\theta_k$;\quad $t\leftarrow0$;\quad $\bar{r}^{\,\ast}\leftarrow-\infty$;\quad $t^{\ast}\leftarrow0$
    \REPEAT
        \STATE $t\leftarrow t+1$;\quad $\theta_{\mathrm{old}}\leftarrow\theta$;\quad draw the next prompt batch $\mathcal{B}_t$ from $\mathcal{D}_{\mathrm{weight}}$
        \STATE collect $\mathcal{S}_t^+$ under $(\theta_{\mathrm{old}},h_k)$
        \STATE ascend $\widehat{J}_{\mathrm{weight}}(\theta;\theta_{\mathrm{old}},h_k)$ over minibatches of $\mathcal{S}_t^+$ (Eq.~\ref{eq:rsft})
        \STATE $\bar{r}_t\leftarrow$ mean verifier reward of the rollouts of the most recent $W$ steps
        \IF{$\bar{r}_t>\bar{r}^{\,\ast}$}
            \STATE $\bar{r}^{\,\ast}\leftarrow\bar{r}_t$;\quad $t^{\ast}\leftarrow t$
        \ENDIF
    \UNTIL{$t>T_{\min}$ \AND $t-t^{\ast}\ge P_{\mathrm{w}}$}
    \STATE $\theta_{k+1}\leftarrow\theta$;\quad $h^{\mathrm{init}}\leftarrow h_k$;\quad $\mathcal{A}_1\leftarrow\{h^{\mathrm{init}}\}$;\quad $\mathcal{E}_1\leftarrow\{e_{h^{\mathrm{init}}}\}$
    \STATE $j\leftarrow1$;\quad $j^{\ast}\leftarrow1$;\quad $J^{\ast}\leftarrow\widehat{J}_{\mathrm{harness}}(\theta_{k+1},h^{\mathrm{init}})$
    \REPEAT
        \STATE $j\leftarrow j+1$;\quad $\mathcal{C}_j\leftarrow\operatorname{Propose}(\mathcal{A}_{j-1},\mathcal{E}_{j-1};M)$
        \STATE evaluate $\widehat{J}_{\mathrm{harness}}(\theta_{k+1},h)$ for each $h\in\mathcal{C}_j$ (Eq.~\ref{eq:mh-score}), yielding artifacts $e_h$
        \STATE $\mathcal{A}_{j}\leftarrow\mathcal{A}_{j-1}\cup\mathcal{C}_j$;\quad $\mathcal{E}_{j}\leftarrow\mathcal{E}_{j-1}\cup\{e_h : h\in\mathcal{C}_j\}$
        \IF{$\underset{h\in\mathcal{C}_j}{\max}\ \widehat{J}_{\mathrm{harness}}(\theta_{k+1},h)>J^{\ast}$}
            \STATE $J^{\ast}\leftarrow\underset{h\in\mathcal{C}_j}{\max}\ \widehat{J}_{\mathrm{harness}}(\theta_{k+1},h)$;\quad $j^{\ast}\leftarrow j$
        \ENDIF
    \UNTIL{$j\ge J_{\min}$ \AND $j-j^{\ast}\ge P_{\mathrm{h}}$}
    \STATE $h_{k+1}\leftarrow\underset{h\in\mathcal{A}_{j}}{\arg\max}\ \widehat{J}_{\mathrm{harness}}(\theta_{k+1},h)$
\ENDFOR
\RETURN the pair $(\theta_k,h_k)$ with the best accuracy on $\mathcal{D}_{\mathrm{test}}$
\end{algorithmic}
\end{algorithm}

\section{Experimental Details}
\label{app:implementation}

\begin{table*}[t]
\caption{Principal experimental settings for the three domains.}
\label{tab:principal-config}
\centering
\small
\renewcommand{\arraystretch}{1.06}
\begin{tabular*}{\textwidth}{@{\extracolsep{\fill}}lccc@{}}
\toprule
Setting & SearchQA & Math & Chess Puzzles \\
\midrule
\rowcolor{gray!12}\multicolumn{4}{c}{\textbf{Models and datasets}} \\
Base model & Qwen3.5-2B & Qwen3.5-2B & Qwen3.5-4B \\
Weight-update training examples & 18,946 & 17,917 & 16,384 \\
Harness-search training examples, $|\mathcal{D}_{\mathrm{harness}}|$ & 256 & 256 & 256 \\
Test examples & 700 & 60 & 256 \\
\addlinespace[2pt]
\rowcolor{gray!12}\multicolumn{4}{c}{\textbf{Weight-only configuration}} \\
Weight-only epoch budget & 4 & 6 & 4 \\
Prompt batch, $|\mathcal{B}_s|$ & 256 & 256 & 256 \\
SFT epochs per accepted trajectory set & 1 & 1 & 1 \\
Learning rate & $10^{-7}$ & $10^{-7}$ & $10^{-7}$ \\
SFT minibatch, $|\mathcal{M}|$ & 64 & 64 & 64 \\
Rollouts per prompt, $G$ & 8 & 8 & 8 \\
\addlinespace[2pt]
\rowcolor{gray!12}\multicolumn{4}{c}{\textbf{Harness-only configuration}} \\
Harness-only iterations, $I$ & 40 & 60 & 40 \\
Proposer & \multicolumn{3}{c}{Claude Opus 4.7} \\
Proposer sessions per harness-search iteration & 1 & 1 & 1 \\
Candidate harnesses per harness-search iteration, $M$ & 3 & 3 & 3 \\
Rollouts per candidate--example pair & 1 & 1 & 1 \\
\addlinespace[2pt]
\rowcolor{gray!12}\multicolumn{4}{c}{\textbf{\method{} and FST configuration}} \\
Train epochs per cycle, $E$ & 0.6 & 0.6 & 0.6 \\
Search iterations per cycle, $I$ & 6 & 6 & 6 \\
Harness search space & \multicolumn{3}{c}{full harness (\method{}); system and user prompts (FST)} \\
Completed cycles & 6 & 9 & 6 \\
\addlinespace[2pt]
\rowcolor{gray!12}\multicolumn{4}{c}{\textbf{Sampling and evaluation}} \\
Test rollouts per example & 8 & 8 & 8 \\
Sampling temperature & 1.0 & 1.0 & 1.0 \\
Top-$p$ & 1.0 & 1.0 & 1.0 \\
Top-$k$ & 20 & 20 & 20 \\
Prompt token limit & 2,048 & 2,048 & 4,096 \\
Response length & 8,192 & 8,192 & 16,384 \\
\bottomrule
\end{tabular*}
\end{table*}

\subsection{Principal Settings}

Table~\ref{tab:principal-config} collects the settings of every reported run. The domains differ only in model size, dataset sizes, total budgets, and context lengths, the longer prompt limit in Chess Puzzles matching its longer board-state observations. Within a domain the four systems draw on the same blocks: weight-only on the first, harness-only on the second, \method{} and FST on both under the per-cycle budgets of the third.

Four points the table does not state. The weight update runs its single SFT epoch over the trajectories accepted in the step that produced them, so no trajectory is revisited later. A harness-search iteration scores every candidate with one rollout per harness-search training example, costing $|\mathcal{D}_{\mathrm{harness}}|\times M$ rollouts in every domain. FST is run with RSFT replacing its reinforcement-learning weight update and MH, restricted to the system and user prompts of $h_0$, replacing its reflective prompt evolution. The weight-only epoch budgets match the cumulative training epochs of the \method{} and FST runs, and the harness-only iteration budgets their cumulative harness-search iterations.

\subsection{Harness-Search Spaces and Base Harness}
\label{app:harness-space}

Section~\ref{sec:experimental-settings} states what the harness search may modify in every domain, and FST is confined to the system and user prompts of $h_0$. Below are each domain's additions to that space and its $h_0$.

\paragraph{SearchQA.}
The harness search may additionally rewrite or expand a model-generated query before retrieval and set retrieval parameters and the ranking of retrieved documents. Each tool call accepts one query, and at most 16 assistant turns are allowed. The initial harness instructs the model to answer factual questions, use search when needed, and place the final answer within \texttt{<answer>} tags; it forwards queries unchanged, returns one passage truncated to 200 tokens, and terminates after two assistant turns or a response without a tool call.

\paragraph{Mathematical Reasoning.}
The harness search may additionally change how Python code is extracted and normalized before execution and how standard output and execution errors are surfaced. Each code-interpreter call accepts one program, and at most 16 assistant turns are allowed. The initial harness has no system prompt and requires a final answer in \verb|\boxed{...}|; it extracts the first code block, inserts a print call for a final expression that is not explicitly printed, returns standard output and error text verbatim, and terminates after two assistant turns or a response without a tool call.

\paragraph{Chess Puzzles.}
The harness search may additionally change how the board observation is presented, how one UCI move is extracted and malformed or ambiguous outputs are handled, how accepted moves and opponent replies are retained or shortened, the retry budgets, and whether the parsed move is validated against the visible legal-move list; it may not generate, substitute, or search for a move. Retry budgets are capped at 10 each and at most 18 assistant turns are allowed. The initial harness asks for exactly one UCI move chosen from the visible position, presents the board and the complete move history without compaction, gives corrective feedback after malformed or illegal output, permits one retry of each kind, and allows nine assistant turns.

\subsection{Binary Verifiers}

Trajectory acceptance in training, harness-candidate scoring, and test evaluation share one binary verifier $R(x,\tau)\in\{0,1\}$ per domain, fixed with its reference answers while $\theta$ and $h$ change.

\paragraph{SearchQA.}
The verifier extracts the answer enclosed by \texttt{<answer>} tags in the final assistant response, returning $R(x,\tau)=0$ without a judge call if parsing fails. Otherwise GPT-5.4-mini, at temperature zero and a 256-token cap, judges the extracted answer against the reference answers under a fixed rubric, and $R(x,\tau)=1$ on a \textsc{correct} judgment. Appendix~\ref{app:search-judge-prompt} reproduces the complete judge prompt.

\paragraph{Mathematical Reasoning.}
Following DAPO's strict boxed-answer verification procedure~\citep{yu2025dapo}, the verifier extracts the final \verb|\boxed{...}| answer. It returns $R(x,\tau)=1$ if the extracted answer matches the ground-truth answer and $R(x,\tau)=0$ if parsing fails or the answers differ.

\paragraph{Chess Puzzles.}
The verifier parses a single UCI move from the model response, checks its legality with \texttt{python-chess}, and compares it with the reference move. Failed parsing, failed legality, or a legal move other than the reference one gives $R(x,\tau)=0$; only a completed reference sequence gives $R(x,\tau)=1$.

\subsubsection{SearchQA LLM-as-a-Judge Prompt}
\label{app:search-judge-prompt}

The SearchQA judge prompt in full. Placeholders are filled per call with the question, the reference answers, and the extracted model answer.

\begin{Verbatim}[
  breaklines=true,
  breakanywhere=true,
  fontsize=\scriptsize,
  frame=single,
  framesep=2mm,
  framerule=0.4pt,
  rulecolor=\color{black!50}
]
You are an impartial judge for a question-answering task.

Mark CORRECT when predicted answer refers to the same entity/value/fact
as any reference. Acceptable:
(1) Surface variation (case, articles, punctuation)
(2) Alias / abbreviation / full-vs-short form ('Pete' = 'Peter')
(3) Added/removed qualifier ('actor X' = 'X')
(4) Less-specific date/place CONTAINED IN reference ('1962' if reference is
    'September 10, 1962'; 'Wellington' if reference is 'Wellington, New Zealand')
(5) Number/unit form ('4' = 'four', '188 acres' = '188 acre')
(6) Different framing of the SAME geographic feature (e.g., 'Gulf of California'
    = 'between Baja California and Sonora' — both describe the same body of water)
(7) Self-correcting answer — if the prediction explores alternatives but commits
    to a final answer that matches the reference, count CORRECT
    (e.g., 'X... actually, Y' where Y matches reference)

Mark INCORRECT for:
(a) Different specific person/place/song/team/company/year
(b) Numeric/date mismatch where neither contains the other
    ('40 miles' vs '35 miles'; '1985' vs '1986')
(c) Opposite or contradictory meaning (winner vs loser; 'no one' vs a named entity)
(d) Different ENTITY TYPE even if related (city is NOT stadium; block is NOT channel;
    episode title is NOT episode number)
(e) Failure to commit to an entity (only explanation/disclaimer)
(f) Listing multiple distinct candidates joined by separators (slashes "/",
    pipes "|", commas, "or", "either…or") — even if one candidate matches
    the reference. A valid answer commits to a single entity; "Beijing /
    Tokyo / Seoul" listing the gold among 2+ alternatives is INCORRECT.
(g) Verbatim passage or document quote — prediction is a multi-sentence
    passage, paragraph, or doc-formatted text (e.g., contains "Doc N (Title:
    ...)" patterns) that contains the reference answer as a substring but
    is not itself a concise answer. A valid answer is the entity/value
    alone, not the surrounding evidence prose.

Question: {question}
Reference answer(s): {gold}
Predicted answer: {prediction}

Respond with exactly one word: CORRECT or INCORRECT.
\end{Verbatim}

\section{Additional Results}
\label{app:additional-results}

\subsection{Main Comparison: Per-Benchmark Results}
\label{app:per-benchmark}

Table~\ref{tab:main-results} breaks the main comparison of Section~\ref{sec:results} down to the seven SearchQA datasets, the two AIME datasets, and the Lichess puzzle test set. \method{} is best on every one, so its domain-level gains are not carried by a subset of benchmarks.

\begin{table*}[!ht]
\caption{Test \meanatk{} accuracy (\%) across domains and test dataset subsets. \method{} uses $(E,I)=(0.6,6)$. Bold denotes the best result in each column.}
\label{tab:main-results}
\centering
\scriptsize
\setlength{\tabcolsep}{1.0pt}
\renewcommand{\arraystretch}{1.08}
\begin{tabular*}{\textwidth}{@{\extracolsep{\fill}}lrrrrrrrr@{\hspace{4pt}}rrr@{\hspace{4pt}}rr@{}}
\toprule
\rowcolor{blue!8}
\multicolumn{1}{c}{\cellcolor{white}} & \multicolumn{8}{c}{\textbf{SearchQA}} & \multicolumn{3}{c}{\textbf{Math}} & \multicolumn{2}{c}{\textbf{Chess Puzzles}} \\
\cmidrule(lr){2-9}\cmidrule(lr){10-12}\cmidrule(l){13-14}
Method & 2Wiki & Bamb. & Hotpot & MuSiQue & NQ & PopQA & Trivia & \cellcolor{gray!12}Avg. & AIME24 & AIME25 & \cellcolor{gray!12}Avg. & Lichess & \cellcolor{gray!12}Avg. \\
\midrule
Harness-only & 26.38 & 36.75 & 38.38 & 13.88 & 46.00 & 40.13 & 66.50 & \cellcolor{gray!12}38.29 & 0.83 & 0.00 & \cellcolor{gray!12}0.42 & 19.82 & \cellcolor{gray!12}19.82 \\
Weight-only & 28.38 & 31.50 & 34.88 & 14.50 & 51.00 & 40.88 & 66.75 & \cellcolor{gray!12}38.27 & 16.25 & 14.58 & \cellcolor{gray!12}15.42 & 22.17 & \cellcolor{gray!12}22.17 \\
FST & 22.25 & 25.00 & 34.25 & 13.63 & 46.88 & 40.62 & 64.75 & \cellcolor{gray!12}35.34 & 15.83 & 20.00 & \cellcolor{gray!12}17.92 & 25.68 & \cellcolor{gray!12}25.68 \\
\method{} & \textbf{35.75} & \textbf{53.00} & \textbf{50.25} & \textbf{22.13} & \textbf{55.50} & \textbf{49.25} & \textbf{72.50} & \cellcolor{gray!12}\textbf{48.34} & \textbf{27.08} & \textbf{22.50} & \cellcolor{gray!12}\textbf{24.79} & \textbf{29.83} & \cellcolor{gray!12}\textbf{29.83} \\
\bottomrule
\end{tabular*}
\end{table*}

\subsection{Schedule Comparison: Best Points}
\label{app:schedule-best}

Table~\ref{tab:schedule-best} tabulates the schedule comparison of Section~\ref{sec:cadence-default}, plotted in Figure~\ref{fig:cadence-ablation}.

\begin{table}[!ht]
\caption{Best test \meanatk{} accuracy (\%) of each run in Figure~\ref{fig:cadence-ablation} and the rollouts consumed up to that point. Bold denotes the best accuracy in each domain and underlining the second best.}
\label{tab:schedule-best}
\centering
\small
\renewcommand{\arraystretch}{1.06}
\begin{tabular*}{\textwidth}{@{\extracolsep{\fill}}lrrrr@{}}
\toprule
\rowcolor{blue!8}
\multicolumn{1}{c}{\cellcolor{white}} & \multicolumn{2}{c}{\textbf{SearchQA}} & \multicolumn{2}{c}{\textbf{Math}} \\
\cmidrule(lr){2-3}\cmidrule(l){4-5}
Run & \meanatk{} & Rollouts & \meanatk{} & Rollouts \\
\midrule
Weight-only & 38.27 & 530,432 & 15.42 & 645,120 \\
Harness-only & 38.29 & 30,720 & 0.42 & 46,080 \\
Stagewise $(E^{\ast},I^{\ast})$ & 43.02 & 604,160 & 15.63 & 893,952 \\
\addlinespace[2pt]
$(E,I)=(0.2,2)$ & 46.63 & 526,336 & 16.67 & 240,128 \\
$(E,I)=(0.2,6)$ & \underline{50.09} & 589,824 & \textbf{28.33} & 599,040 \\
$(E,I)=(0.6,2)$ & 48.84 & 378,880 & 17.92 & 582,656 \\
$(E,I)=(0.6,6)$ & 48.34 & 382,464 & 24.79 & 720,384 \\
$(E,I)=(1.0,10)$ & 45.93 & 294,400 & 24.79 & 510,464 \\
\addlinespace[2pt]
Adaptive \method{} & \textbf{52.82} & 295,424 & \underline{26.46} & 689,152 \\
\bottomrule
\end{tabular*}
\end{table}

\ifarxiv\else
\subsection{Adaptive \method{}: Realized Schedules}
\label{app:realized-schedules}

\begin{figure}[!ht]
\centering
\includegraphics[width=\textwidth]{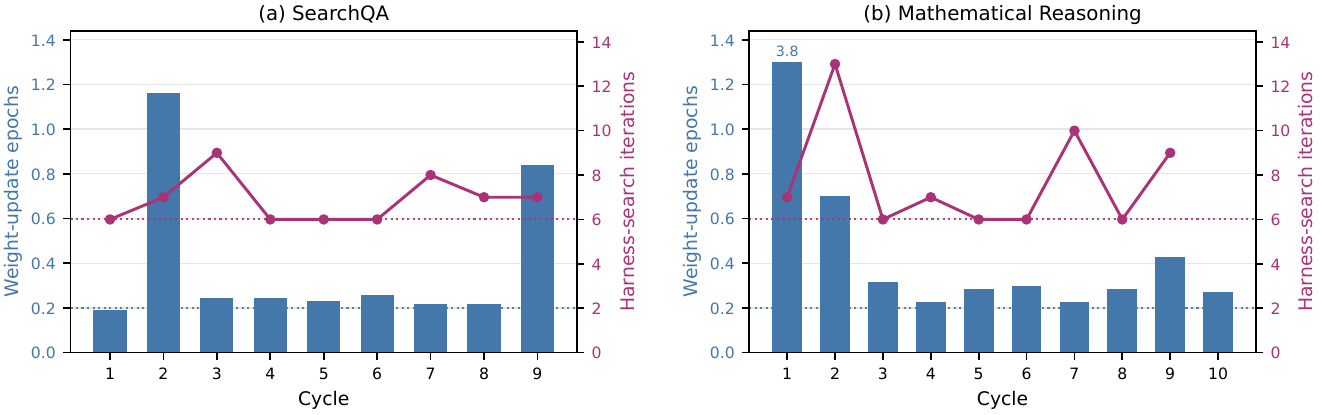}
\caption{Realized per-cycle budgets chosen by the patience rule of adaptive \method{} in (a) SearchQA and (b) Mathematical Reasoning: weight-update epochs (bars, left axis) and harness-search iterations (dots, right axis). Dotted horizontal lines mark the minimum phase lengths (0.2 epoch, $I=6$).}
\label{fig:adaptive-schedule}
\end{figure}

Figure~\ref{fig:adaptive-schedule} shows the budget the patience rule realized in each cycle of the two runs evaluated in Section~\ref{sec:adaptive-schedule}. Most cycles stop at or just above the floors of 0.2 weight-update epochs and 6 harness-search iterations, so extensions are occasional rather than systematic. Harness-search iterations vary more than weight-update epochs, extending whenever a late proposal sets a new best training score.
\fi

% The trajectory blocks cannot be split across pages, so most pages here
% are underfull; collect the slack at the foot of the page instead of
% stretching it into the body.
\raggedbottom
\section{Trajectory Examples}
\label{app:trajectories}

For each domain we show one test item that the base model under $h_0$ answers incorrectly and that \method{} with $(E,I)=(0.6,6)$ answers correctly, with the two trajectories side by side. Message roles are colour-coded: system prompt, user prompt, model response, tool call, tool response, and, in Chess Puzzles, the environment's reply. Long spans are trimmed where marked.

\subsection{SearchQA}
\label{app:traj-search}

The question is \emph{Where was the husband of Octavie Coudreau born?}, whose reference answers are \texttt{Sonnac} and \texttt{Sonnac, Charente-Maritime}. The base pair scores 0/8, \method{} 8/8.

\paragraph{Base model with the base harness $h_0$ (incorrect).}
The harness allows two assistant turns and returns one passage. The first retrieval names the husband but not his birthplace; the model searches again, and the rollout ends with no \texttt{<answer>} tag.

\begin{trajsys}
\begin{Verbatim}[breaklines=true,fontsize=\scriptsize]
# Tools

You have access to the following functions:

<tools>
{"type": "function", "function": {"name": "search", "description": "Searches the web for relevant information based on a single query. Returns the top-k passages for that one query.\n\nMirrors `verl/tools/search_tool.py:SearchTool.execute

[... trimmed ...]

- If there is no function call available, answer the question like normal with your current knowledge and do not tell the user about function calls
</IMPORTANT>

Answer factual questions using search when needed.
\end{Verbatim}
\end{trajsys}

\begin{trajusr}
\begin{Verbatim}[breaklines=true,fontsize=\scriptsize]
Question: Where was the husband of Octavie Coudreau born?
Use the search tool if needed. Put the final answer only inside <answer>...</answer>.
\end{Verbatim}
\end{trajusr}

\begin{trajasst}
\begin{Verbatim}[breaklines=true,fontsize=\scriptsize]
I need to find information about Octavie Coudreau, specifically her husband's birthplace. Let me search for this information.
</think>
\end{Verbatim}
\end{trajasst}

\begin{trajcall}
\begin{Verbatim}[breaklines=true,fontsize=\scriptsize]
<tool_call>
<function=search>
<parameter=query_list>
["Octavie Coudreau husband birthplace"]
</parameter>
<parameter=topk>
5
</parameter>
<parameter=max_doc_tokens>
200
</parameter>
</function>
</tool_call>
\end{Verbatim}
\end{trajcall}

\begin{trajresp}
\begin{Verbatim}[breaklines=true,fontsize=\scriptsize]
<tool_response>
{"result": "Doc 1 (Title: \"Octavie Coudreau\")\nOctavie Coudreau Marie Octavie Coudreau (\"\"née\"\" Renard; 1867–1938) was a French explorer from Anais, Charente, and author of several books on French Guiana and northern Brazil. In 1899, the Brazilian states of Pará and Amazonas hired Coudreau to explore and chart the Amazon region. Her husband was the French explorer and geographer Henri Coudreau.

[... trimmed ...]
\end{Verbatim}
\end{trajresp}

\begin{trajasst}
\begin{Verbatim}[breaklines=true,fontsize=\scriptsize]
<think>
The search result mentions that Octavie Coudreau's husband was Henri Coudreau, but it doesn't state where Henri Coudreau was born. Let me try a more specific search to find information about Henri Coudreau's birthplace.
</think>
\end{Verbatim}
\end{trajasst}

\begin{trajcall}
\begin{Verbatim}[breaklines=true,fontsize=\scriptsize]
<tool_call>
<function=search>
<parameter=query_list>
["Henri Coudreau birthplace"]
</parameter>
<parameter=topk>
3
</parameter>
<parameter=max_doc_tokens>
200
</parameter>
</function>
</tool_call>
\end{Verbatim}
\end{trajcall}

\paragraph{\method{} with $(E,I)=(0.6,6)$ (correct).}
Everything that differs on the prompt side is written by the searched harness. Its system prompt grants three search calls and four assistant turns, against the two turns of $h_0$, and requires at least one retrieval. Its user prompt appends a checklist: the passage must state the answer verbatim, and the answer must be one specific value. Between turns the harness inserts a state line listing the titles retrieved so far. The model searches again when the first passage misses the birthplace, then commits.

\begin{trajsys}
\begin{Verbatim}[breaklines=true,fontsize=\scriptsize]
[... trimmed ...]

s using search when needed. You have a budget of up to three search calls and up to four assistant turns total. You MUST consult the retriever at least once before your final answer; do not rely on parametric memory alone. If your first search does not fully answer the question, use a follow-up search that targets a MORE SPECIFIC missing fact (a different entity, a linked concept, or a narrower time/place) — do not repeat the same query. On or before the fourth turn, commit to your final answer inside <answer>...</answer> (a single name, date, number, or short noun phrase).
\end{Verbatim}
\end{trajsys}

\begin{trajusr}
\begin{Verbatim}[breaklines=true,fontsize=\scriptsize]
Question: Where was the husband of Octavie Coudreau born?

Use the search tool if needed. After each retrieval, before producing <answer>, verify:
  (1) Does a retrieved passage state the specific answer verbatim (a name, date, or number that directly answers the question)? If not, search again with a DIFFERENT phrasing or clue.
  (2) If the question has multiple constraints (e.g. "X who did A and B"), does your candidate answer satisfy EVERY constraint, not just one?
  (3) Your answer must be a single specific value (name, date, number, or short noun phrase) — not a category, list, or explanation.
Put the final answer only inside <answer>...</answer>.
\end{Verbatim}
\end{trajusr}

\begin{trajasst}
\begin{Verbatim}[breaklines=true,fontsize=\scriptsize]
The user is asking a question: "Where was the husband of Octavie Coudreau born?"

I need to find information about Octavie Coudreau and then find out where her husband was born.

First, I need to find information about Octavie Coudreau. I should search for information about her to find out who her husband is, and then find out where that husband was born.
</think>
\end{Verbatim}
\end{trajasst}

\begin{trajcall}
\begin{Verbatim}[breaklines=true,fontsize=\scriptsize]
<tool_call>
<function=search>
<parameter=query_list>
["Octavie Coudreau", "Octavie Coudreau husband", "Octavie Coudreau"]
</parameter>
<parameter=topk>
3
</parameter>
<parameter=max_doc_tokens>
200
</parameter>
</function>
</tool_call>
\end{Verbatim}
\end{trajcall}

\begin{trajresp}
\begin{Verbatim}[breaklines=true,fontsize=\scriptsize]
<tool_response>
{"result": "Doc 1 (Title: \"Octavie Coudreau\")\nOctavie Coudreau Marie Octavie Coudreau (\"\"née\"\" Renard; 1867–1938) was a French explorer from Anais, Charente, and author of several books on French Guiana and northern Brazil. In 1899, the Brazilian states of Pará and Amazonas hi

[... trimmed ...]
\end{Verbatim}
\end{trajresp}

\begin{trajusr}
\begin{Verbatim}[breaklines=true,fontsize=\scriptsize]
Titles retrieved so far: Octavie Coudreau. You have used 1 of 4 assistant turns; up to 2 more searches are possible. If any title in the list above already contains the answer, commit now inside <answer>...</answer>. Otherwise, your next search MUST target a NEW entity or attribute not already represented in the list — do NOT re-query a title above.
\end{Verbatim}
\end{trajusr}

\begin{trajcall}
\begin{Verbatim}[breaklines=true,fontsize=\scriptsize]
<tool_call>
<function=search>
<parameter=query_list>
["Henri Coudreau birth place", "Henri Coudreau birth"]
</parameter>
<parameter=topk>
3
</parameter>
<parameter=max_doc_tokens>
200
</parameter>
</function>
</tool_call>
\end{Verbatim}
\end{trajcall}

\begin{trajresp}
\begin{Verbatim}[breaklines=true,fontsize=\scriptsize]
<tool_response>
{"result": "Doc 1 (Title: \"Henri Coudreau\")\nHenri Coudreau Henri Anatole Coudreau (6 May 1859 Sonnac – 10 November 1899, State of Pará, Brazil) was a French professor of history and geography, explorer and geographer of French Guiana and the tributaries of the Amazon. At the time

[... trimmed ...]
\end{Verbatim}
\end{trajresp}

\begin{trajasst}
\begin{Verbatim}[breaklines=true,fontsize=\scriptsize]
and the search result explicitly states he was born in **Sonnac** (6 May 1859).

<answer>Sonnac</answer>
</think>

The search results identify her husband as **Henri Coudreau** and the text explicitly states he was born in **Sonnac**.

<answer>Sonnac</answer>
\end{Verbatim}
\end{trajasst}

\subsection{Mathematical Reasoning}
\label{app:traj-math}

The problem is an AIME 2025 counting question with reference answer \texttt{279}. The base pair scores 0/8, \method{} 8/8.

\paragraph{Base model with the base harness $h_0$ (incorrect).}
The response never calls the code interpreter. It reasons in prose for 20,035 characters, reaches the token limit and is truncated mid-sentence, so the verifier finds no \verb|\boxed{...}| answer to extract.

\begin{trajsys}
\begin{Verbatim}[breaklines=true,fontsize=\scriptsize]
# Tools

You have access to the following functions:

<tools>
{"type": "function", "function": {"name": "code_interpreter", "description": "A tool for executing code.", "parameters": {"properties": {"code": {"type": "string", "description": "The code to execute."}}, "required": ["code"], "type": "object"}}}
</tools>

If you choose to call a function ONLY reply in the following format with NO suffix:

<tool_call>
<function=example_function_name>
<parameter=example_parameter_1>
value_1
</parameter>
<parameter=example_parameter_2>
This is the value for the second parameter
that can span
multiple lines
</parameter>
</function>
</tool_call>

<IMPORTANT>
Reminder:
- Function calls MUST follow the specified format: an inner <function=...></function> block must be nested within <tool_call></tool_call> XML tags
- Required parameters MUST be specified
- You may provide optional reasoning for your function call in natural language BEFORE the function call, but NOT after
- If there is no function call available, answer the question like normal with your current knowledge and do not tell the user about function calls
</IMPORTANT>
\end{Verbatim}
\end{trajsys}

\begin{trajusr}
\begin{Verbatim}[breaklines=true,fontsize=\scriptsize]
There are $8!=40320$ eight-digit positive integers that use each of the digits $1,2,3,4,5,6,7,8$ exactly once. Let $N$ be the number of these integers that are divisible by 22. Find the difference between $N$ and 2025.
The answer format must be: \boxed{'The final answer goes here.'}
\end{Verbatim}
\end{trajusr}

\begin{trajasst}
\begin{Verbatim}[breaklines=true,fontsize=\scriptsize]
The user wants to find the number of permutations of the digits $\{1, 2, 3, 4, 5, 6, 7, 8\}$ that form an 8-digit integer divisible by 22. Then subtract 2025 from this number.

Let the digits be $x_1, x_2, x_3, x_4, x_5, x_6, x_7, x_8$ in some order.
The digits are distinct and form the sequence $1, 2, 3, 4, 5, 6, 7, 8$.
The number must be divisible by 22.
$22 = 2 \times 11$. So the number must be divisible by 2 and by 11.

Condition 1: Divisible by 2.
This means the last digit (x_8) must be even. The even digits a

[... trimmed ...]

Positions are $1, 2, 3, 4, 5, 6, 7, 8$.
We need to place digits $D_O$ into positions $1, 3, 5, 7$ and $D_E$ into positions $2, 4, 6, 8$.
Wait, the positions are distinct.
So if we have a fixed set of digits for positions $1,3,5,7$, say $\{a, b, c, d\}$, there are $4!$ ways
\end{Verbatim}
\end{trajasst}

\paragraph{\method{} with $(E,I)=(0.6,6)$ (correct).}
The harness leaves the opening prompt untouched; the system and user messages are byte-identical to the base pair's. What it changes is the turn budget, seven assistant turns against two, and what follows execution: the tool response carries the interpreter's output and then a reminder to return a settled result as \verb|\boxed{...}| rather than as another call. The switch from prose to a single code call is therefore attributable to the weight updates, which is what Section~\ref{sec:task-dependent-roles} means by model-dominant. The response is 3,149 characters and ends in a boxed answer.

\begin{trajsys}
\begin{Verbatim}[breaklines=true,fontsize=\scriptsize]
# Tools

You have access to the following functions:

<tools>
{"type": "function", "function": {"name": "code_interpreter", "description": "A tool for executing code.", "parameters": {"properties": {"code": {"type": "string", "description": "The code to execute."}}, "required": ["code"], "type": "object"}}}
</tools>

If you choose to call a function ONLY reply in the following format with NO suffix:

<tool_call>
<function=example_function_name>
<parameter=example_parameter_1>
value_1
</parameter>
<parameter=example_parameter_2>
This is the value for the second parameter
that can span
multiple lines
</parameter>
</function>
</tool_call>

<IMPORTANT>
Reminder:
- Function calls MUST follow the specified format: an inner <function=...></function> block must be nested within <tool_call></tool_call> XML tags
- Required parameters MUST be specified
- You may provide optional reasoning for your function call in natural language BEFORE the function call, but NOT after
- If there is no function call available, answer the question like normal with your current knowledge and do not tell the user about function calls
</IMPORTANT>
\end{Verbatim}
\end{trajsys}

\begin{trajusr}
\begin{Verbatim}[breaklines=true,fontsize=\scriptsize]
There are $8!=40320$ eight-digit positive integers that use each of the digits $1,2,3,4,5,6,7,8$ exactly once. Let $N$ be the number of these integers that are divisible by 22. Find the difference between $N$ and 2025.
The answer format must be: \boxed{'The final answer goes here.'}
\end{Verbatim}
\end{trajusr}

\begin{trajasst}
\begin{Verbatim}[breaklines=true,fontsize=\scriptsize]
We have a problem: "There are $8!=40320$ eight-digit positive integers that use each of the digits $1,2,3,4,5,6,7,8$ exactly once. Let $N$ be the number of these integers that are divisible by 22. Find the difference between $N$ and 2025.

So we need to count how many 8-digit numbers formed by digits 1-8 (distinct) that are divisible by 22. Since 22 = 2*11. So number must be divisible by 2 (i.e., even) and by 11.

The digits a

[... trimmed ...]

So just brute force.

We have digits 1-8. 8! = 40320. That's manageable.

Let's brute force.

We can write Python to count.

Then output final answer.

We must output answer as a boxed value.

Okay, let's compute.
</think>
\end{Verbatim}
\end{trajasst}

\begin{trajcall}
\begin{Verbatim}[breaklines=true,fontsize=\scriptsize]
<tool_call>
<function=code_interpreter>
<parameter=code>
import itertools

def count_divisible_by_22():
    digits = [1,2,3,4,5,6,7,8]
    import itertools
    # generate permutations
    import itertools
    digits = [1,2,3,4,5,6,7,8]
    from itertools import permutations
    count=0
    for p in itertools.permutations(digits, 8):
        # form number
        # digits are 1-8 so no leading zero issue
        # convert to int
        n = int("".join(map(str, p)))
        # check 22
        if n % 22 == 0:
            count+=1
    return count

count = count_divisible_by_22()
count
</parameter>
</function>
</tool_call>
\end{Verbatim}
\end{trajcall}

\begin{trajresp}
\begin{Verbatim}[breaklines=true,fontsize=\scriptsize]
<tool_response>
2304

[reminder] If the answer is now clear from this output, your next reply should be the final answer in \boxed{<answer>} -- no more tool calls are required. Only call code_interpreter again if you spotted a specific numerical error.
</tool_response>
\end{Verbatim}
\end{trajresp}

\begin{trajasst}
\begin{Verbatim}[breaklines=true,fontsize=\scriptsize]
<think>
We found 2304 numbers. That's the number of 8-digit integers using digits 1-8 exactly once divisible by 22.

Now we need to find N = 2304.

Find difference between N and 2025.

2304 - 2025 = 279.

So answer = 2304 - 2025 = 279.

Let's double-check logic: we need 8-digit numbers using digits 1-8 exactly once. That's 8! = 40320 permutations. We counted those divisible by 22. That's correct.

Thus N = 2304.

N - 2025 = 27

[... trimmed ...]

A number is divisible by 22 if it's divisible by 2 and 11.
- Divisible by 2: last digit must be even. Digits available: 1,2,3,4,5,6,7,8 -> even digits: 2,4,6,8.
- Divisible by 11: alternating sum of digits must be divisible by 11.

We can brute force to find the count N.

\boxed{279}
\end{Verbatim}
\end{trajasst}

\subsection{Chess Puzzles}
\label{app:traj-chess}

\begin{figure}[!ht]
\centering
\includegraphics[width=0.66\textwidth]{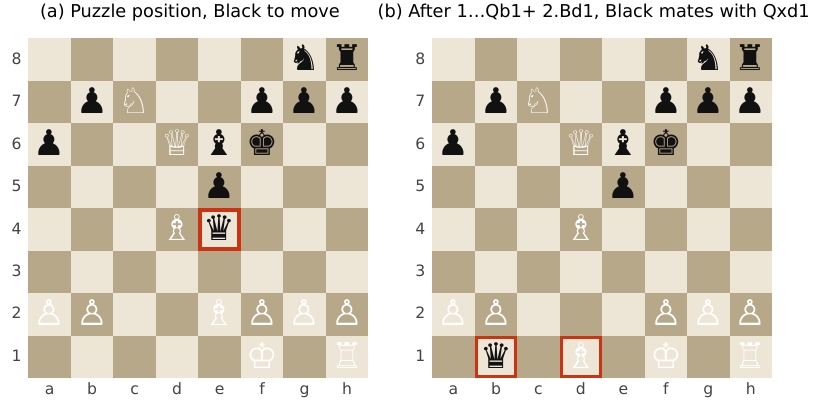}
\caption{The Chess Puzzles example. \textbf{(a)} the puzzle position, with the
black queen on e4 outlined. \textbf{(b)} after $1\ldots$Qb1+ 2.Bd1, where
Qxd1 mates.}
\label{fig:chess-example-board}
\end{figure}

The puzzle is a mate in two for Black, shown in Figure~\ref{fig:chess-example-board}. The base pair scores 0/8, \method{} 8/8.

\paragraph{Base model with the base harness $h_0$ (incorrect).}
The base harness states the position as a FEN string and an ASCII board, lists the legal moves flat, and asks for one move. The model never commits: it deliberates to the 8,129-token cap and is cut off mid-word, emitting no \texttt{<move>} tag while quoting 21 bracketed UCI tokens from the list, so the verifier reports \texttt{\_\_ambiguous\_multiple\_moves\_\_}. The harness offers a retry, shown below, but the budget is spent and no second turn follows. This is typical: 2,025 of the 2,048 base rollouts stop at the same cap, averaging 8,112 assistant tokens against 764 under \method{}.

\begin{trajsys}
\begin{Verbatim}[breaklines=true,fontsize=\scriptsize]
You solve chess tactics by choosing one legal move at a time.
Use only the board position and legal moves shown in the prompt. Do not claim
hidden engine analysis or external knowledge. Return exactly one move.
\end{Verbatim}
\end{trajsys}

\begin{trajusr}
\begin{Verbatim}[breaklines=true,fontsize=\scriptsize]
Solve the current chess puzzle position.

Task: solve the chess puzzle one move at a time.
Side to move: Black
FEN: 6nr/1pN2ppp/p2Qbk2/4p3/3Bq3/8/PP2BPPP/5K1R b - - 0 19
Board:
8 . . . . . . n r
7 . p N . . p p p
6 p . . Q b k . .
5 . . . . p . . .
4 . . . B q . . .
3 . . . . . . . .
2 P P . . B P P P
1 . . . . . K . R
  a b c d e f g h
Puzzle move number: 1
Legal moves:
- [a6a5] SAN=a5
- [b7b5] SAN=b5
- [b7b6] SAN=b6
- [e4b1] SAN=Qb1+
- [e4c2] SAN=Qc2
- [e4c6] SAN=Qc6
- [e4d3] SAN=Qd3
- [e4d4] SAN=Qxd4
- [e4d5] SAN=Qd5
- [e4e2] SAN=Qxe2+
- [e4e3] SAN=Qe3
- [e4f3] SAN=Qf3
- [e4f4] SAN=Qf4
- [e4f5] SAN=Qf5
- [e4g2] SAN=Qxg2+
- [e4g4] SAN=Qg4
- [e4g6] SAN=Qg6
- [e4h4] SAN=Qh4
- [e5d4] SAN=exd4
- [f6f5] SAN=Kf5
- [f6g5] SAN=Kg5
- [f6g6] SAN=Kg6
- [g7g5] SAN=g5
- [g7g6] SAN=g6
- [g8e7] SAN=Ne7
- [g8h6] SAN=Nh6
- [h7h5] SAN=h5
- [h7h6] SAN=h6

Return exactly one move as <move>uci</move>, for example <move>e2e4</move>.
Do not include explanation.
\end{Verbatim}
\end{trajusr}

\begin{trajasst}
\begin{Verbatim}[breaklines=true,fontsize=\scriptsize]
The user wants me to solve a chess puzzle.
I need to determine the best move for Black given the current position.

**Position Analysis:**
FEN: `6nr/1pN2ppp/p2Qbk2/4p3/3Bq3/8/PP2BPPP/5K1R b - - 0 19`
Side to move: Black

**Board Breakdown:**
- Black pieces:
    - Q on e4
    - N on g8
    - B on b4 (Wait, looking at FEN, 'b' on rank 6 is a king. Wait, let's parse FEN carefully)

[... trimmed ...]

Black Bishop on b7? `1pN2ppp` `p` on b7.
            Black Knight on c7? No.
            Black Bishop on e6? Maybe.
            Is there any protection?
            If `g6h7` pawn exists.
            Wait, `7...`
            Black King on f6.

            Wait, I
\end{Verbatim}
\end{trajasst}

\begin{trajenv}
\begin{Verbatim}[breaklines=true,fontsize=\scriptsize]
The verifier could not parse exactly one UCI move from "__ambiguous_multiple_moves__". Reply with exactly one move, e.g. <move>e2e4</move>, and no explanation.
\end{Verbatim}
\end{trajenv}

\begin{trajusr}
\begin{Verbatim}[breaklines=true,fontsize=\scriptsize]
[... trimmed ...]

[f6g5] SAN=Kg5
- [f6g6] SAN=Kg6
- [g7g5] SAN=g5
- [g7g6] SAN=g6
- [g8e7] SAN=Ne7
- [g8h6] SAN=Nh6
- [h7h5] SAN=h5
- [h7h6] SAN=h6

Feedback from verifier:
The verifier could not parse exactly one UCI move from "__ambiguous_multiple_moves__". Reply with exactly one move, e.g. <move>e2e4</move>, and no explanation.

Return exactly one move as <move>uci</move>, for example <move>e2e4</move>.
Do not include explanation.
\end{Verbatim}
\end{trajusr}

\paragraph{\method{} with $(E,I)=(0.6,6)$ (correct).}
The prompt below is the searched harness's own work. It parses a piece list from the board and splits it by side, then groups the legal moves by origin square and annotates them \texttt{(mate)}, \texttt{(check)} and \texttt{(capture)}. Each turn restates the moves accepted so far and the opponent's last reply, so the position need not be carried in the reasoning; the environment itself returns only the short line between the two prompts. The harness also fixes how a move is read off a reply: bracketed \texttt{[uci]} tokens inside \texttt{<think>} are scratch notes, and the last \texttt{<move>} tag outside it is the commitment. That rule disposes of the failure above. The model plays Qb1+, sees Bd1 interpose, and mates with Qxd1, in 828 generated tokens.

\begin{trajsys}
\begin{Verbatim}[breaklines=true,fontsize=\scriptsize]
You solve chess tactics one move at a time.
Reply with exactly this two-part structure:
<think>at most three short lines of reasoning about the visible legal moves</think>
<move>PUT_UCI_HERE</move>

Rules for the <move> tag:
- Replace PUT_UCI_HERE with a UCI string copied verbatim from a bracketed
  [uci] entry in the legal moves list, for example <move>e2e4</move> or
  <move>b7f7</move>.
- DO NOT emit the literal pl

[... trimmed ...]

- If you emit multiple <move> tags outside <think> (for example because you
  revised your pick after a second look), the LAST <move> tag is treated as
  your final commitment. Earlier <move> tags are scratch.
Do not claim hidden engine analysis or external knowledge.
\end{Verbatim}
\end{trajsys}

\begin{trajusr}
\begin{Verbatim}[breaklines=true,fontsize=\scriptsize]
Solve the current chess puzzle position.

Side to move: Black
Board:
8 . . . . . . n r
7 . p N . . p p p
6 p . . Q b k . .
5 . . . . p . . .
4 . . . B q . . .
3 . . . . . . . .
2 P P . . B P P P
1 . . . . . K . R
  a b c d e f g h

Piece coordinates (parsed from the ASCII board above, K=king Q=queen R=rook B=bishop N=knight P=pawn):
Your pieces (you are BLACK, lowercase on the board): Ng8, Rh8, Pb7, Pf7, Pg7, Ph7, Pa6, Be6, Kf6, Pe5, Qe4
Opponent pieces (WHITE, uppercase on the board): Nc7, Qd6, Bd4, Pa2, Pb2, Be2, Pf2, Pg2, Ph2, Kf1, Rh1

Legal moves (28 total, grouped by origin square):

From a6 (P on a6) - 1 move(s):
  - [a6a5] SAN=a5

From b7 (P on b7) - 2 move(s):
  - [b7b5] SAN=b5
  - [b7b6] SAN=b6

From e4 (Q on e4) - 15 move(s):
  - [e4b1] SAN=Qb1+  (check)
  - [e4c2] SAN=Qc2
  - [e4c6] SAN=Qc6
  - [e4d3] SAN=Qd3
  - [e4d4] SAN=Qxd4  (capture)
  - [e4d5] SAN=Qd5
  - [e4e2] SAN=Qxe2+  (check, capture)
  - [e4e3] SAN=Qe3
  - [e4f3] SAN=Qf3
  - [e4f4] SAN=Qf4
  - [e4f5] SAN=Qf5
  - [e4g2] SAN=Qxg2+  (check, capture)
  - [e4g4] SAN=Qg4
  - [e4g6] SAN=Qg6
  - [e4h4] SAN=Qh4

From

[... trimmed ...]

- If you cannot fit reasoning, skip <think> and reply only with <move>UCI</move>.
- Any bracketed [uci] tokens inside <think> are treated as scratch notes, not
  as your commitment.
- If you emit multiple <move> tags outside <think> (because you revised your
  pick), the LAST <move> tag outside <think> is treated as your final
  commitment. Earlier <move> tags are scratch.
\end{Verbatim}
\end{trajusr}

\begin{trajasst}
\begin{Verbatim}[breaklines=true,fontsize=\scriptsize]
The position shows Black to move. The legal moves list includes a checkmate move [e4b1] SAN=Qb1+ which is a check, but I need to verify if it's actually checkmate or just check. Looking at the legal moves provided, [e4b1] is listed as a legal move for Black's queen to b1. However, I need to find the best move. The list shows 28 total legal moves. The most forcing move would be checkmate if available, or check if it l

[... trimmed ...]

</think>

Black's queen can deliver check to the white king on b1.
This is the only check move available in the legal moves list.
<move>e4b1</move>
\end{Verbatim}
\end{trajasst}

\begin{trajenv}
\begin{Verbatim}[breaklines=true,fontsize=\scriptsize]
Correct. Opponent replied e2d1 (Bd1). Continue with the next move.
\end{Verbatim}
\end{trajenv}

\begin{trajusr}
\begin{Verbatim}[breaklines=true,fontsize=\scriptsize]
Solve the current chess puzzle position.

Side to move: Black
Board:
8 . . . . . . n r
7 . p N . . p p p
6 p . . Q b k . .
5 . . . . p . . .
4 . . . B . . . .
3 . . . . . . . .
2 P P . . . P P P
1 . q . B . K . R
  a b c d e f g h
Last opponent reply: e2d1
Your accepted moves so far: e4b1

Piece coordinates (parsed from the ASCII board above, K=king Q=queen R=rook B=bishop N=knight P=pawn):
Your pieces (you are BLACK, lowercase on the board): Ng8, Rh8, Pb7, Pf7, Pg7, Ph7, Pa6, Be6, Kf6, Pe5, Qb1
Opponent pieces (WHITE, uppercase on the board): Nc7, Qd6, Bd4, Pa2, Pb2, Pf2, Pg2, Ph2, Bd1, Kf1, Rh1

Legal moves (23 total, grouped by origin square):

From a6 (P on a6) - 1 move(s):
  - [a6a5] SAN=a5

From b1 (Q on b1) - 10 move(s):
  - [b1a1] SAN=Qa1
  - [b1a2] SAN=Qxa2  (capture)
  - [b1b2] SAN=Qxb2  (capture)
  - [b1c1] SAN=Qc1
  - [b1c2] SAN=Qc2
  - [b1d1] SAN=Qxd1#  (mate, capture)
  - [b1d3] SAN=Qd3+  (check)
  - [b1e4] SAN=Qe4
  - [b1f5] SAN=Qf5
  - [b1g6] SAN=Qg6

From b7 (P on b7) - 2 move(s

[... trimmed ...]

pick), the LAST <move> tag outside <think> is treated as your final
  commitment. Earlier <move> tags are scratch.
\end{Verbatim}
\end{trajusr}

\begin{trajasst}
\begin{Verbatim}[breaklines=true,fontsize=\scriptsize]
<think>
The board position is critical. Black to move. The legal moves include [b1d1] which is Qxd1# (checkmate). This is the immediate winning move for Black if the pieces are set up as described, but I must verify if this is indeed a legal move for Black.
The list says "From b1 (Q on b1) ... - [b1d1] SAN=Qxd1# (mate, capture)".
Wait, the prompt says "Your pieces (you are BLACK, lowercase on the board)".
But the leg

[... trimmed ...]

Therefore, the move is b1d1.

<move>b1d1</move>
</think>

Black to move and delivers immediate checkmate with the queen on b1 capturing on d1.
This move ends the game immediately.
The legal moves list confirms [b1d1] is mate.

<move>b1d1</move>
\end{Verbatim}
\end{trajasst}

\begin{trajenv}
\begin{Verbatim}[breaklines=true,fontsize=\scriptsize]
Correct. Puzzle solved.
\end{Verbatim}
\end{trajenv}

\end{document}